 \documentclass[5p]{elsarticle}
 \usepackage{lineno,hyperref}
 \modulolinenumbers[20]

\usepackage{doi}
\usepackage[utf8]{inputenc} 
\usepackage[english]{babel}
\usepackage{changepage}
\usepackage{xargs}
\usepackage{siunitx}
\DeclareSIUnit[number-unit-product = {}]{\inchQ}{\,''}
\DeclareSIUnit[number-unit-product = {\thinspace}]{\inch}{in}
\DeclareSIUnit[number-unit-product = {\thinspace}]{\px}{px}
\usepackage{mathtools,bm}
\usepackage{amsmath,amsfonts,amssymb}

\DeclareMathOperator*{\argmax}{arg\,max}
\usepackage{csquotes}
\usepackage{graphicx}
\usepackage[pdftex,dvipsnames]{xcolor}
\usepackage{layouts}
\usepackage{stfloats}
\usepackage[font={small}]{caption}

\biboptions{sort&compress}

\newcommand{\LR}{{L\mkern-3mu/\mkern-4muR}}
\newcommand{\xy}{{x\mkern-3mu/\mkern-3muy}}

\journal{Neural Networks}

\biboptions{authoryear}
\let\cite\citep 

\begin{document}

\begin{frontmatter}

  \title{
    Exploitation of Image Statistics with Sparse Coding \\
    in the Case of Stereo Vision
  }

  \author[tue,dai]{Gerrit A. Ecke\corref{correspondingauthor}}
  \ead{gerrit.ecke@daimler.com}
  \author[tue]{Harald M. Papp}
  \author[tue]{Hanspeter A. Mallot}

  \address[tue]{Cognitive Neuroscience Unit,
    Department of Biology,
    University of T\"ubingen,
    Auf der Morgenstelle 28,
    72076 T\"ubingen }
  \address[dai]{Mercedes-Benz AG,
    Leibnizstraße 2,
    71032 B\"oblingen }

  \cortext[correspondingauthor]{Corresponding author}

  \begin{abstract}
    The sparse coding algorithm has served as a model
    for early processing in mammalian vision.
    It has been assumed that the brain uses sparse coding
    to exploit statistical properties of the sensory stream.
    We hypothesize that sparse coding discovers patterns from the data set,
    which can be used to estimate a set of stimulus parameters 
    by simple readout.
    In this study, we chose a model of stereo vision to test our hypothesis.
    We used the Locally Competitive Algorithm (LCA),
    followed by a na\"ive Bayes classifier,
    to infer stereo disparity.
    From the results we report three observations.
    First, disparity inference was successful
    with this naturalistic processing pipeline.
    Second, an expanded, highly redundant representation is required
    to robustly identify the input patterns.
    Third, the inference error can be predicted
    from the number of active coefficients in the LCA representation.
    We conclude that sparse coding can generate a suitable general representation
    for subsequent inference tasks.
  \end{abstract}


\end{frontmatter}

\linenumbers

\section{Introduction}
\label{sec:introduction}
Among neural coding principles that have been proposed over time,
sparse coding has a long-standing and successful history
in explaining properties of neuronal circuitry.
The firing rates of visual cortical neurons
follow a sparse regime \cite{rolls_sparseness_1995,baddeley_responses_1997,froudarakis_population_2014}
and several algorithms that model this premise
predict receptive fields of visual cortex neurons
quite accurately
(see \citet{olshausen_emergence_1996,
  hyvarinen_image_1998,
  ringach_spatial_2002,
  rehn_network_2007,
  hunter_distribution_2015}).

It is not straight forward to understand
why sparse representations evolved in the brain.
A possible explanation is based on the assumption
that the neuronal code exploits statistical properties of the sensory input \cite{simoncelli_natural_2001, barlow_exploitation_2001}.
Sparse coding represents the sensory input with a low number of specialized units
that make the higher order, redundant components of a signal explicit \cite{field_relations_1987,bethge_factorial_2006,eichhorn_natural_2009}.
This specialization is reminiscent of Barlow's concept of specialist units,
or \emph{cardinal cells},
with a selectivity intermediate between that of
concrete \emph{pontifical neurons} or \emph{grandmother cells}
and that of a typical distributed representation \cite{barlow_single_1972,barlow_redundancy_2001}.
Cardinal cells could represent faces, objects, or, as Barlow puts it,
\emph{``a pattern of external events of the order of complexity
  of the events symbolized by a word''} \cite{barlow_single_1972}.

The sensory visual stream contains evidence
for external events of various degrees of abstraction
that are relevant for an animal to detect.
Examples are the occurrence of a specific texture,
an object that can be assigned to a category,
or subtle cues,
like signs of social interaction.
We hypothesize that sparse coding supports the exploitation of sensory statistics
by the formation of \emph{cardinal cells}
that make a subset of these external events accessible with a \emph{simple readout} method.

Sparse coding transforms the sensory stream \(\bm{x}\)
into a representation \(\bm{h} = H(\bm{x})\).
\(\bm{h}\) is the vector of activities of a set of \emph{cardinal cells},
with an intermediate selectivity to external events \(\{y_i\}\).
We further assume that the selectivity of cardinal cells 
allows us to detect the occurrence of elements of \(\{y_i\}\)
with a simple processing step \({\hat y} = Y(\bm{h})\).
For this \emph{simple readout}
we chose a na\"ive Bayes classifier
\begin{equation}
  \hat{y} = \argmax_i \, P(y_i) \prod_{k=1}^K  P \left( h_k \,|\, y_i \right)
\end{equation}
with uniform prior \(P(y_i)\).
It selects the external event \(y_i\) that most likely occurred in the sensory stream,
based on evidence from the \(K\) elements \(h_k\) in~\(\bm{h}\).

The readout \({Y}(\bm{h})\) assumes independence of the elements of \(\bm{h}\).
Sparse coding belongs to the class of independent component analysis algorithms (ICA)
that aim to extract basis vectors which are statistically independent \cite{hyvarinen_independent_2000}.
Note that, in the case of image data,
the independence of basis vectors obtained by standard ICA algorithms
is known to be strongly violated \cite{bethge_factorial_2006,eichhorn_natural_2009}.
Interestingly, classifiers that assume independence often yield surprisingly good results,
even though existing dependencies between variables are omitted \cite{hand_idiots_2001,zhang_exploring_2005,kuncheva_optimality_2006,kupervasser_mysterious_2014}.

It is unclear how to identify the set of external events
that is accessible with this simple readout.
However, assuming that the striate cortex forms a representation akin to sparse coding,
we can use physiological evidence to identify candidates.
For our evaluation we therefore adopt stereo vision,
which is an early detection task.
Indeed, we can compare our results with a large body of literature
that is concerned with stereo vision in biological systems.

The contributions of this paper are:
(\emph{i.})
a characterization of stereo kernels learned with the Locally Competitive Algorithm (LCA),
and their associated tuning to disparity and surface orientation
in comparison to physiological findings,
(\emph{ii.})
an evaluation of disparity inference
with simple readout from the LCA representation,
subject to sparsity load and overcompleteness,
(\emph{iii.})
a method to predict the inference error,
based on the number of active coefficients in the LCA representation.

\section{Related work} \label{sec:related}

\subsection{Compact vs.\ expanded codes}
Barlow reasoned in his efficient coding hypothesis
that an efficient code,
stripped by its redundancies,
makes information more accessible,
just as reducing the size of a haystack
simplifies the task of finding needles \cite{barlow_sensory_1959}.
He later extended this view by arguing that,
in such a compact representation,
interference between several, simultaneously occurring events
might impair their separability \cite{barlow_redundancy_2001}.
\citet{gardner-medwin_limits_2001} hypothesized that
event retrieval from a population code is optimal
when overlap between the neurons that correspond to each event is minimal.
They tested their assumption by linking each of a number of hypothetical events
to a fixed, random subset of binary neurons within a population.
Overlap then was subject to two degrees of freedom:
the number of neurons that, on average, corresponded to an event,
and the total number of neurons in the population.
Results indicated minor (but evident) impact of mean neural activity,
but strong impact of population size on the readout error.
Their findings suggest that an expanded,
exceedingly redundant representation
provides an optimal basis for event retrieval.
An encoding with the sparse coding algorithm
transforms the sensory stream
into such an expanded, redundant representation \cite{field_what_1994}.
Moreover,
Gardner-Medwin and Barlow varied mean activity and population size,
which are also parameters of the sparse coding algorithm.
The population size corresponds to the dimensionality,
which is usually several times overcomplete,
and activity can be adjusted by the sparsity load of the optimization.
We report how varying these parameters effects disparity inference
in Sec.~\ref{sec:res:overcompleteness} and Sec.~\ref{sec:res:sparsity}.

\subsection{Sparse coding and pattern recognition}
\citet{rigamonti_are_2011} examined the sparse coding algorithm
as the first processing step in an image classification pipeline.
They found that the features extracted by sparse coding were superior to handcrafted features
even when they were used as a simple convolutional filter bank.
They also evaluated the classification error as a function of sparsity penalty.
No substantial improvement over convolutional processing was found.
Better classification performance was monotonically linked to \emph{lower} sparsity penalty.
\citet{kurkova_sparsity_2018} found the opposite:
better classification performance with larger sparsity penalty,
however with the very specialized MNIST dataset.
Also employing the MNIST dataset for evaluation,
\citet{kurkova_strategies_2018} imposed sparsity on a perceptron-like feed forward network
by adjusting neural thresholds,
and similarly obtained a positive correlation
between high sparsity penalty and classification performance.

\subsection{Sparse coding and stereo vision} \label{sec:stereoICA}
A considerable amount of work on stereo vision with unsupervised learning methods
was carried out in the context of independent component analysis (ICA).
\citet{hoyer_independent_2000} applied ICA to color- and stereo images
and received disparity tuned Gabor-like basis vectors.
Left and right basis vectors were similar,
but varied in position and phase, as well as in the the degree of ocular dominance.
\citet{hunter_distribution_2015} performed a thorough analysis
of ICA stereo basis vectors,
obtained from a database carefully adjusted to the human visual system.
The most notable difference to physiological data
was two modes in the difference of phase
between left and right basis vectors.
The two modes were at zero
and at \(\pi\) radians phase-shift,
i.e., with opposite polarity.
This finding might be related to a model from \citet{li_efficient_1994},
who derived kernels for correlated and anticorrelated
left and right stereo half-images.

\citet{lonini_robust_2013} found that a sparse representation
can be learned altogether with vergence control.
They reasoned that the angular orientation of both eyes
has significant impact on achievable optimality of the representation.
In their model, vergence control was a function of the global distribution of disparities.
This is in line with psychophysical experiments with humans,
which fits a population coding model
that minimizes overall disparity energy in the two half-images \cite{mallot_disparity-evoked_1996}.

\citet{lundquist_sparse_2016,lundquist_sparse_2017} used stereo sparse coding,
followed by a classifier,
for depth inference,
as well as for object detection.
Their model outperformed others in the case of limited labeled training data.
They concluded that the competition inherent in sparse coding
requires elements to match associated contextual cues.
\citet{timofte_sparse_2015} tackled the associated problem of optic flow detection
with a model which performed competitive to state of the art algorithms.

\subsection{Stereo vision in biological systems} \label{sec:stereoBiology}
In the visual cortex of mammals,
most cells in foveal striate and prestriate cortex show binocular interaction \cite{
  hubel_receptive_1962,
  hubel_stereoscopic_1970,
  poggio_binocular_1977,
  levay_ocular_1978,
  guillemot_binocular_1993,
  tanabe_suppressive_2011,
  hubel_binocular_2015}.
Binocular simple cells are similar to kernels obtained with sparse coding or ICA.
They best respond to Gabor-like binocular stimuli,
with slight differences in position and phase \cite{anzai_neural_1999}.
V1 receptive fields are, however, more variant,
with a tendency to appear more blob-like,
with fewer sinusoidal sub-fields \cite{ringach_spatial_2002}.
Binocular complex cells are more generally tuned to disparity than binocular simple cells,
irrespective of position and polarity of the stimulus within the receptive field.
In the standard model,
complex cells are driven
by a quadrature pair of Gabor-like monocular simple cells \cite{ohzawa_stereoscopic_1990}.

Robust disparity inference requires further processing.
Two constraints are crucial for the recovery of depth.
First, each location in one stereo half-image
corresponds to at most one location in the other half-image.
Second, depth varies smoothly in general \citet{marr_cooperative_1976,marr_computational_1979}.
The constraints hold well,
with the exception of strong local violation at the edges of objects.
Optimization for both constraints yields the disparity of corresponding image locations.
With epipolar geometry and with known distance of the eyes,
disparity can be used to calculate depth \cite{hartley_multiple_2004}.
\citet{read_sensors_2007} presented a model
which relates the correspondence problem to differences in position- and phase
of the receptive fields of binocular simple cells.
Equally shaped Gabor filters that only vary in position
are the best match for the corresponding structures in both half-images,
whereas phase-shift Gabor filters carry the information to detect false matches.
\citet{goncalves_what_2017} showed that a simple readout of disparity from simple cells
incorporates this information.

We assume that a representation built by sparse coding provides a generalizing,
yet limited basis for a range of pattern detection tasks.
In order to test this assumption,
we experimented with the detection
of other characteristics of spatial layout than disparity.
Psychophysical findings indicate
that many more cues than point disparities
contribute to a complete understanding of spatial layout.
Examples include orientations of lines,
light intensity differences,
disparate specular highlights,
and monocular occlusions.
For an overview of geometrical and global aspects of stereopsis see \citet{mallot_stereopsis_1999}.
Neurons in caudal intraparietal area
were shown to be selective for first order depth,
i.e., for specific surface tilt- and slant angles,
and neurons in the temporal sulcus
were shown to be selective for second order depth,
i.e., for concave and convex curvature \cite{orban_extraction_2011}.
Responses of such neurons
were highly specific and robust against texturing and other orders of depth.
We therefore decided to test the sparse coding representation for first order depth selectivity.
For an overview of physiological aspects of higher order visual processing of 3D-shape in the brain
see \citet{orban_higher_2008}.

\section{Databases}
Analyses of this paper rely on four databases.
The virtual vergence database was used for LCA optimization,
the disparity database and the naturalistic scene database for disparity inference,
and the surface orientation database was used to characterize LCA selectivity to surface orientation.

\subsection{Virtual vergence database}

\begin{figure*}[b]
  \includegraphics[width=\textwidth]{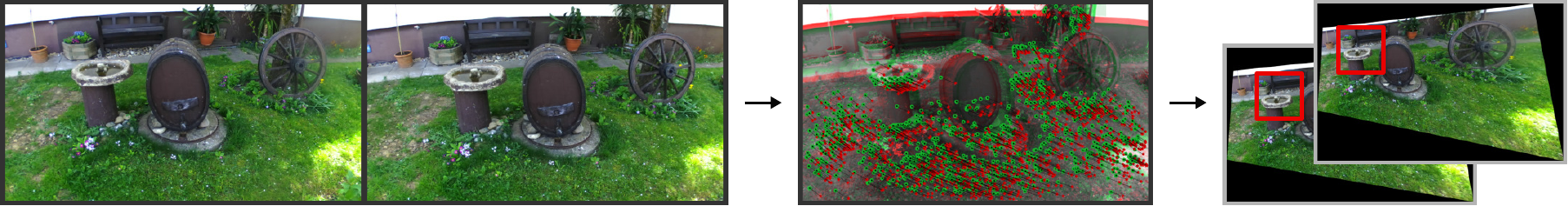}
  \caption{
    The virtual vergence database
    was created from images captured with a ZED stereo camera
    with parallel principal axes.
      {\bf a)}~Example image (view cross-eyed).
      {\bf b)}~Corresponding image points (SURF features)
    in the left and right half-images (red and green respectively, anaglyph image)
    were automatically matched and selected.
      {\bf c)}~Example stereo image with virtual vergence,
    created from {\bf a}~by correcting for radial distortion and applying homography transformation.
    Red frames indicate the extend of the final database images.
  }
  \label{fig:DatabaseCreationWorkflow}
\end{figure*}

We captured images around T\"ubingen, Germany, with a ZED stereo camera\footnote{
  \url{https://www.stereolabs.com/}}.
The camera was equipped with two \SI{1/3}{\inchQ} sensors,
fixed at \SI{120}{\mm} distance,
with parallel principal axes.
The fields of view covered \ang{76}(H)~\(\times\)~\ang{47}(V),
with a resolution of \num{2208 x 1242}\,px.
With \(f_{\xy}=1400\)\,px,
the central angular resolution was \(\sim\)\num{0.04}~degrees.
Note that the angular resolution 
of the final images we used in the subsequent processing steps was \(\sim\)\num{0.08}~degrees,
as described in detail below.
Image data were stored lossless as \num{24}~bit png-files
after automated brightness and gamma correction.
In total, \num{1081} pairs of pictures were taken,
from which \num{222} were captured inside rooms,
\num{480} showed man made outdoors structures
and the remaining \num{379} comprised natural scenes.

In order to obtain a database with vergence towards corresponding locations,
images with several virtual fixations were created from each captured stereo image
(see Fig.~\ref{fig:DatabaseCreationWorkflow}).
SURF-features \cite{bay_speeded-up_2008} from left and right half-images were brute-force matched
by the metric distance between the feature vectors.
Only sufficiently similar matches below a threshold were selected
and outliers with respect to the epipolar constraint were excluded.

For a given stereo image pair, 
a virtual fixation of any point in the image
can be calculated by homographic transformation \cite{hartley_multiple_2004}.
Because the transformation assumes a pinhole camera,
the images were first corrected for radial distortion.
The pixel positions \(\bm{x}\) of these rectified images 
were then shifted to their new positions \(\bm{x}'\).
Assuming a rotation around the camera nodal point,
the shift of each pixel was calculated with
\begin{equation}
  \bm{x}' = KRK^{-1}\bm{x} \, ,
\end{equation}
where \(K\) is the camera matrix and \(R\) is the matrix 
that describes the rotation of the camera.
We used the camera calibration app 
from the MATLAB computer vision toolbox
to estimate the camera matrix \(K\).
The virtual rotation of each camera was determined, according to Listing's law,
as a rotation around the axis \(\bm{u}\)
that is parallel to the image plane
and perpendicular to the vector \(\bm{s} - \bm{p}\)
between the matched SURF-feature location \(\bm{s}\) 
and the principle point of the camera \(\bm{p}\) in the image.
Therefore, the rotation axis was calculated as
\begin{equation}
  \bm{u} = 
  \begin{pmatrix}
    - (s_x - p_x) \\ s_y - p_y \\ 0
  \end{pmatrix}  \, .
\end{equation}
The value of the rotation angle was calculated as
\begin{equation}
  \Theta = \arctan{\frac{\lVert \bm{s} - \bm{p} \rVert}{f_{x/y}}} \, .
\end{equation}
With the normalized vector \(\hat{\bm{u}}  = \bm{u} /  \lVert \bm{u} \rVert \),
the rotation matrix was then obtained by calculating 
\begin{equation}
  R = 
  \begin{bmatrix}
    \scriptstyle
    \cos{\Theta} + \hat{u}_x^2 (1 - \cos{\Theta}) & 
    \scriptstyle
    \hat{u}_x \hat{u}_y (1-\cos{\Theta}) & 
    \scriptstyle
    \hat{u}_y \sin{\Theta} \\
    \scriptstyle
    \hat{u}_x \hat{u}_y (1-\cos{\Theta}) & 
    \scriptstyle
    \cos{\Theta} + \hat{u}_y^2 (1-\cos{\Theta}) & 
    \scriptstyle
    -\hat{u}_x \sin{\Theta} \\
    \scriptstyle
    -\hat{u}_y \sin{\Theta} & 
    \scriptstyle
    \hat{u}_x \sin{\Theta} & 
    \scriptstyle
    \cos{\Theta}
  \end{bmatrix} \! .
\end{equation}

\begin{figure}
  \includegraphics[width=\columnwidth]{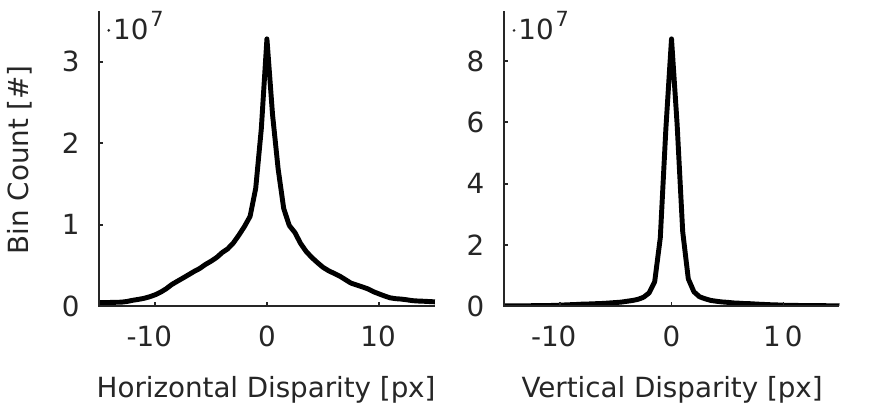}
  \caption{
    Distribution of horizontal and vertical disparities
    in the virtual vergence database.
    The histograms show the distribution
    of \num{3.3e8} randomly drawn data points (bin widths \num{0.5}\,px).
    Disparities were clustered around zero with high kurtosis (\(k_x=29.9\), \(k_y=118.1\)).
  }
  \label{fig:DistributionOfDisparity_virtverg}
\end{figure}

In order to keep local image statistics intact,
we discarded images in which the virtual camera rotation angles exceeded \num{20}~degrees.
Pixel values were mapped to the new pixel raster
and downscaled to half the original resolution with bicubic interpolation.
The angular resolution of the final images was therefore \(\sim\)\num{0.08}~degrees.
They were cropped to \num{256 x 256}\,px,
centered at the respective principal points.
In total, the virtual vergence database consisted of \num{72991} images.
We extracted the distribution of disparities contained in the database
with FlowNet~2.0 \cite{ilg_flownet_2017};
results are shown in Fig.~\ref{fig:DistributionOfDisparity_virtverg}.

\subsection{Disparity database} \label{sec:stereoDatabases}
The disparity database contained stereo images,
where each right stereo half-image was a shifted version of the left half-image.
Images were collected from the same set
that was used to create the virtual vergence database.
It consisted of disparities in the range of \(d_x\),~\(d_y=-6\),~\ldots,~\(6\)\,px,
rasterized by \num{0.5}\,px in both dimensions.
These were processed by cropping out \num{512 x 512}\,px sized pairs,
randomly positioned in the original images,
with \(2\,d_x\) and \(2\,d_y\)\,px left-to-right offset.
Next, they were downscaled to half the original resolution with bilinear filtering.
We obtained \num{500 x 25 x 25} image pairs, each with a resolution of \num{256 x 256}\,px.
From these \num{500} images per stimulus,
\num{490} were used for training
and the remaining \num{10} were used for testing.
We used convolutional LCA from \citet{schultz_replicating_2014}
and accumulated data points over the feature maps (see Sec.~\ref{sec:LCA} and Sec.~\ref{sec:inference}).
After discarding the margins,
each feature map yielded \num{784} data points,
which amounts to a total number of \num{384160} data points
per disparity and kernel for training
and a total number of \num{7840} data points per disparity and kernel for testing.

\subsection{Naturalistic scene database}
We used the publicly available Genua Pesto database \cite{canessa_dataset_2017},
which contains two rendered 3d-scenes
with vergence towards common fixation points.
We used one of these scenes,
the ground truth disparity
and the right half-image of which are shown in Fig.~\ref{fig:GenuaResults}a and~b.

\subsection{Surface orientation database}
The surface orientation database
contained stereo images of surfaces,
textured with images
from the same set that was used to create the virtual vergence database.
With Blender\footnote{\url{https://blender.org}},
two virtual cameras, with a \ang{11.8} field of view,
were placed \SI{7}{\centi\meter} apart
and oriented towards the surface.
The distance from the mid point between the two camera nodes
to the central point of the surface was \SI{1}{\meter}.
The cameras were oriented
so that the principal axes pierced the center of the surface,
mimicking ocular vergence.
We created stereo half-images
for every combination
of \num{36} tilt angles \(\varphi\) and \num{6} slant angles \(\alpha\) with respect to a fronto-parallel plane.
Tilt angles \(\varphi\) were equally spaced by \ang{10},
and slant angles were set to \(\alpha = 6^{\circ}\), \(24.3^{\circ}\), \(38.2^{\circ}\), \(48.2^{\circ}\), \(55.2^{\circ}\).
They were chosen
so that each step increased disparity of a horizontally slanted surface by \num{1}\,px,
assessed at \num{10}\,px horizontal distance from the center.
We additionally included the images with a fronto-parallel plane \(\alpha = 0^{\circ}\).
Per stimulus, we generated a training set with \(L=10050\) images,
and an additional test set with \num{1000} images,
both with \num{256 x 256}\,px resolution.

\section{Modeling the visual processing pipeline}
\begin{figure*}[t]
  \includegraphics[width=\textwidth]{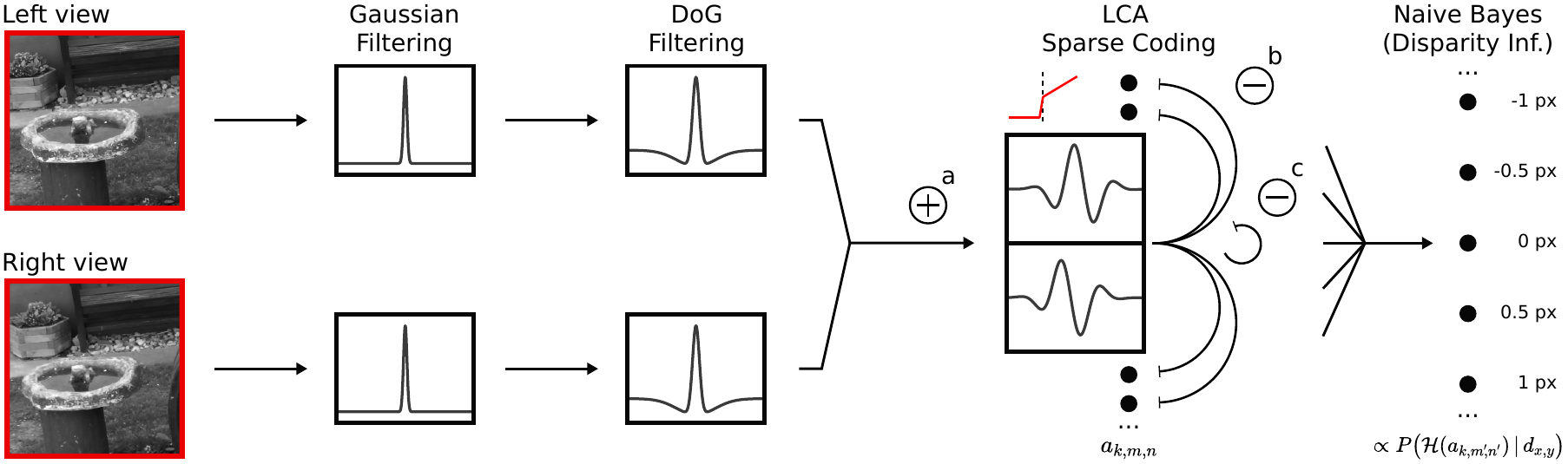}
  \caption{
    Schematic processing pipeline.
    Left and right half-images from the virtual vergence database
    were first pre-processed by a convolution with a Gaussian
    and subsequent difference-of-Gaussians filtering.
    In neural network notion,
    processing with the Locally Competitive Algorithm (LCA)
    is equivalent to a recurrent network.
    It is driven by excitatory feed forward connections,
    with learned weights that are usually Gabor-like (labeled a),
    competition through mutual lateral inhibition,
    with connection strengths proportional to the pairwise similarity of the feed-forward weights (b)
    and self inhibition or leaky integration (c).
    A na\"ive Bayes classifier was used for simple readout.
    It is equivalent to a simple feed-forward network,
    with weights proportional to the LCA neurons' log-probability
    of being active in the presence of a stimulus,
    and an additional winner-take-all mechanism.
  }
  \label{fig:ProcessingPipeline}
\end{figure*}

We built a simplified, naturalistic processing pipeline
that mimics the mammalian visual system.
For an illustration, see Fig.~\ref{fig:ProcessingPipeline}.
Processing started from two horizontally separated eyes,
with vergence towards a common fixation point.
Visual sensory data underwent retinal pre-processing
and were propagated to the model's sub-unit resembling V1,
where a sparse representation was established.
Finally, a na\"ive Bayes classifier was used for simple readout.
If not acknowledged otherwise, implementation was carried out in MATLAB\footnote{
  MATLAB Release 2018a, The MathWorks, Inc., Natick, Massachusetts, United States.}.
The retina model and the LCA sparse coding
were implemented in PetaVision\footnote{
  \url{https://petavision.github.io/}}.

\subsection{Retinal processing}
Retinal processing was modeled in two steps.
First, each image was smoothed by Gaussian filtering (\(\sigma = 0.5\)\,px).
Then, mimicking receptive fields with center-surround organization,
images were convolved with a difference-of-Gaussians filter
(DoG, inner Gaussian: \(\sigma = 1\)\,px,
outer Gaussian: \(\sigma = 5.5\))\,px.
For each Gaussian kernel,
weights were normalized
so that the integral was equal to \(1\).
Before propagation to the LCA sparse coding layer,
each image was mean-centered
and rescaled to a common \( \ell_2 \)-norm.

\subsection{Establishing a sparse representation} \label{sec:LCA}
In order to model V1, we used the locally competitive algorithm (LCA),
introduced by \citet{rozell_sparse_2008},
extended to convolutional LCA by \citet{schultz_replicating_2014}
(see also \citet{zeiler_deconvolutional_2010} and \citet{lundquist_sparse_2016}).
Here, we provide a short summary of the algorithm.
Sparse coding is the optimization of an error function
that consists of two terms:
a reconstruction term for reversibility,
and a penalty which encourages sparsity \cite{olshausen_emergence_1996}.
In the case of stereo sparse coding,
where the inputs were left and right stereo half-images \( \bm{I}_{L} \) and \( \bm{I}_{R} \),
reconstruction was approximated by the convolutions
\begin{equation}
  \bm{I}_{L}
  \approx
  \sum_{k=1}^K {\bm{\Phi}_{{L,k}}} \ast \bm{A}_k
  \;\;\; \text{and} \;\;\;
  \bm{I}_{R}
  \approx
  \sum_{k=1}^K {\bm{\Phi}_{{R,k}}} \ast \bm{A}_k \; .
\end{equation}
\( \Phi_{L} = { \left\{ {\bm{\Phi}_{L,k}} \right\} }_{k=1}^K \)
and \( \Phi_{R} = { \left\{ {\bm{\Phi}_{R,k}} \right\} }_{k=1}^K \)
were sets of left and right half-kernels.
Both half-kernels were convolved with a common corresponding feature map
from the set \( A = { \left\{ \bm{A}_k \right\} }_{k=1}^K \),
so that the reconstruction of the left and the right stereo half-image was coupled.
We jointly normalized the left and right half-image by their \(\ell_2\)-norm,
which enabled learning of monocular dominant kernels.
The particular error function for stereo sparse coding was
\begin{equation} \label{eq:errorFct}
  E = \;
  \frac{1}{2}
  \left(
  \big\|
  R(\bm{I}_L,{\Phi_L},{A})
  \big\|_2^2
  +
  \big\|
  R(\bm{I}_R,{\Phi_R},{A})
  \big\|_2^2
  \right)
  \;+\;
  S({A}) \, ,
\end{equation}
with the reconstruction term
\begin{equation}
  R(\bm{I}_\LR,{\Phi_\LR},{A}) \, = \,
  \bm{I}_\LR \, - \, \sum_{k=1}^K {\bm{\Phi}_{{\LR},k} \ast \bm{A}_k} \, .
\end{equation}
For standard sparse coding,
the sparsity penalty \(S(A)\)
is the \(\ell_1\)-norm of the coefficients of \(A\) \cite{tibshirani_regression_1996,olshausen_emergence_1996}.
The LCA penalizes the number of super-threshold coefficients, given a threshold \(\lambda\).
With convolutional feature map dimensions \(M \times N\),
and with coefficients \(a_{k,m,n}\) of \(\bm{A}_k\),
the sparsity penalty was
\begin{equation}
  S({A}) = \sum_{k,m,n}  {\cal H}(a_{k,m,n} - \lambda) \, ,
\end{equation}
where \({\cal H}(x) = 1\) if \(x>0\) and \({\cal H}(x) = 0\) otherwise.
Note that this formulation requires the activity to be restricted to \(a_{k,m,n} \geq 0 \),
which is convenient in neural network notion.
Optimization of Eq.~\ref{eq:errorFct} for kernels \(\Phi_\LR\),
as well as activity in \(A\),
was obtained by the gradient descent procedure
described by \citet{rozell_sparse_2008} and \citet{schultz_replicating_2014}.

We set the kernel size to \num{16 x 16}\,px
and the stride of the convolutions to \num{8}\,px,
so that \(k \times 2 \times 2\) elements \(k\), \(m\), \(n\)
contributed to the reconstruction of single image pixels.
We obtained five sets with \(K=\)~\numlist{85; 128; 384; 1024; 2048} kernels respectively,
which constituted \numlist{0.66; 1; 3; 8; 16} times overcomplete representations.
\(\lambda\) was set to \num{0.1} for all models at learning time and was only varied at test time.

\subsection{Simple readout} \label{sec:inference}

\begin{figure*}[t]
  \includegraphics[width=\textwidth]{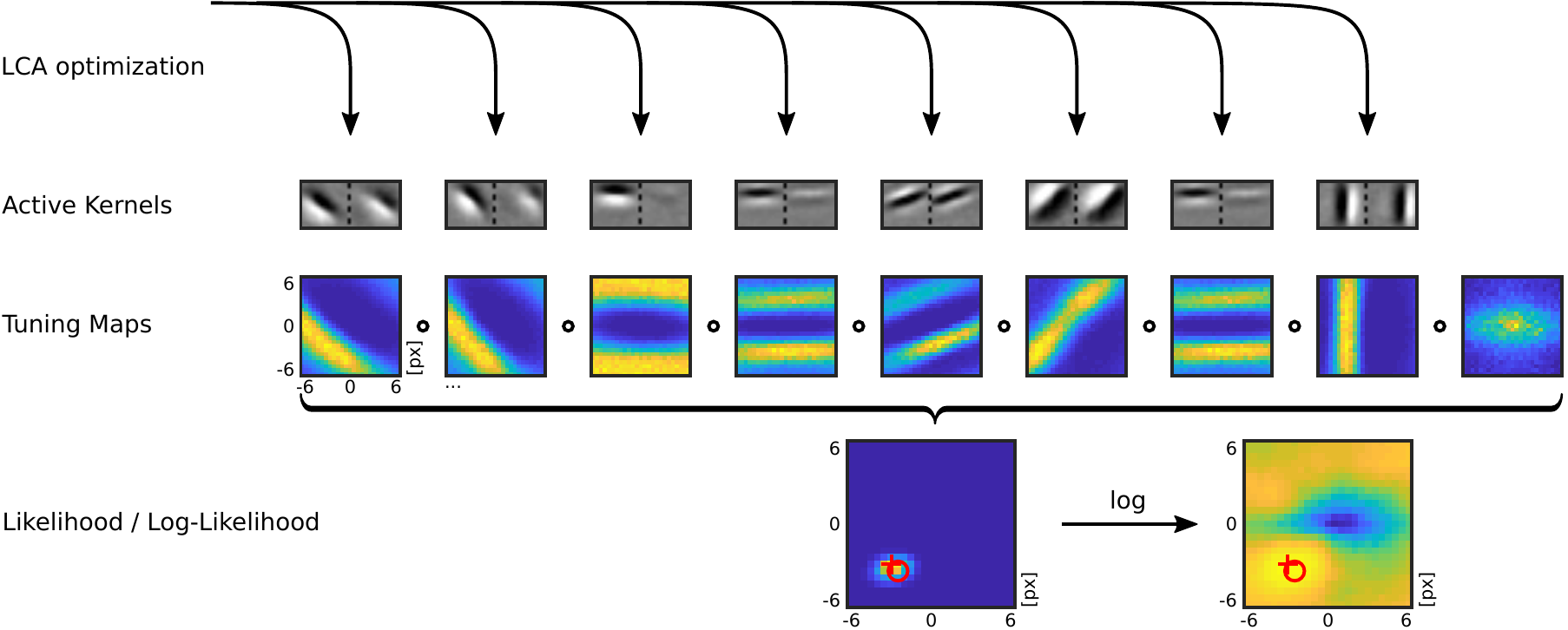}
  \caption{
    Example of disparity inference with the na\"ive Bayes classifier
    from a single image presentation
    with disparity \(d_x = -2.5\)\,px, \(d_y = -4\)\,px.
    LCA optimization results in a sparse set of active coefficients.
    Each binocular kernel \({\bm{\Phi}_{\LR,k}}\) (grayscale)
    is associated with a tuning map (arbitrary units).
    The tuning maps display the probability of the corresponding coefficient \(a_{k,m'\!\!,n'}\)
    to be in an active state,
    as a function of the evaluated range of x- and y-disparities.
    Bottom row: disparity likelihood / log-likelihood (prior omitted).
    The likelihood map is the Hadamard product of all \(8\) tuning maps associated with active coefficients,
    and all inverted tuning maps of non-active units (accumulated, outmost right).
    The true disparity is indicated by a red circle
    and the mode of the distribution is indicated by a red cross.
  }
  \label{fig:ExampleInference}
\end{figure*}

The readout was based on representations
obtained by running the LCA procedure on stereo images,
but with learning of \(\bm{\Phi}_\LR\) turned off.
The kernels were obtained
from the previous LCA optimization on the virtual vergence database.
An example of disparity readout
is visualized in Fig.~\ref{fig:ExampleInference}.
With each image presentation,
the set of feature maps \(A\) was used for inference.
After settled optimization, the coefficients \(a_{k,m,n}\) of all feature maps \(\bm{A}_k\) were set to binary states
by applying \( {\cal H}(a_{k,m,n})\),
with \({\cal H}(x) = 1\) if \(x>0\) and \({\cal H}(x) = 0\) otherwise.
Disparity was inferred
based on the \num{2 x 2} coefficients \(a_{k,m'\!\!,n'}\)
of feature maps \(\bm{A}_k\),
which is the extend of all coefficients that include a single pixel in their receptive fields.
Surface orientation tuning was examined
based on a larger \num{7 x 7} region around the central fixation point.

Each category \(y_i\) in \(\bm{y}\) is represented by a unique two-dimensional parameter combination:
horizontal and vertical disparity \(d_x\) and \(d_y\),
and surface tilt- and slant angles \(\varphi\) and \(\alpha\).
At each image location,
they can be estimated by selecting \(\hat{y}=y_i\)
for some \(i\) that is most probable.
Assuming independence of the coefficients,
estimates were calculated by applying a na\"ive Bayes classifier with\footnote{
  In practice, inference was calculated equivalently
  with the logarithmic form
  \(\hat{y} = \argmax_i \;\log P(y_i) + \sum_{k,m'\!\!,n'} \; \log \; P \left( {\cal H}(a_{k,m'\!\!,n'}) \,|\, y_i \right) \, \).
}
\begin{equation} \label{eq:naiveBayes}
  \hat{y} = \argmax_i \, P(y_i) \prod_{k,m'\!\!,n'}  P \left( {\cal H}(a_{k,m'\!\!,n'}) \,|\, y_i \right) \, .
\end{equation}
We omitted the priors \(P(y_i)\),
even though a strongly non-uniform distribution of disparities is apparent in natural image data,
as can be seen in Fig.~\ref{fig:DistributionOfDisparity_virtverg}.

Because elements \( {\cal H}(a_{k,m'\!\!,n'}) \) were restricted to two states,
the probabilities of being in one of these states,
\(P \left( {\cal H}(a_{k,m'\!\!,n'})=1 \,|\, y_i \right) \)
and \(1 -P \left( {\cal H} (a_{k,m'\!\!,n'})=1 \,|\, y_i \right)  \)
were determined experimentally by calculating
the arithmetic mean.
In the case of stereo disparity,
we assumed that the probabilities were linked to each of the kernels \({\bm{\Phi}_{\LR,k}}\)
and invariant with respect to image feature location.
Therefore, probes were accumulated over the feature maps of size \(M \times N\),
as well as over the whole training set of size \(L\) by calculating
\begin{equation} \label{eq:DisparityTuning}
  P \left( {\cal H}(a_{k,m'\!\!,n'})=1 \,|\, y_i \right)
  \approx
  \frac{1}{L M N} \sum_{l} \sum_{m,n} ({\cal H}(a_{k,m,n}))_l \, .
\end{equation}
For inference with Eq.~\ref{eq:naiveBayes},
the same probability estimate of one kernel
was used for all \num{2 x 2} locations \(m'\!\!,n'\).

In contrast, we assumed that probabilities differ with respect to image location in the case of surface orientation.
We reasoned that kernels are mainly disparity tuned
and that the orientation of a surface may be detected by the pattern of disparities within a local range.
We therefore calculated probability estimates independently for all \num{7 x 7} locations \(m,n\) with
\begin{equation} \label{eq:OrientationTuning}
  P \left( {\cal H}(a_{k,m,n})=1 \,|\, y_i \right)
  \approx
  \frac{1}{L} \sum_{l} ({\cal H}(a_{k,m,n}))_l \, .
\end{equation}

Probability estimates vary smoothly with respect to the parameter combinations \(d_x\) and \(d_y\)
as well as for \(\varphi\) and \(\alpha\) and therefore constitute ``tuning maps''.
In the case of surface orientation estimation we exploited this local continuity
and smoothed out noise with two dimensional Savitzky-Golay filtering \cite{savitzky_smoothing_1964},
with a polynomial of degree \num{3}, and with \num{5}\,px width in both dimensions.

\subsection{Linking the processing pipeline to biological vision} \label{sec:LinkToBiology}
With this study, we hope to contribute to a better understanding of biological vision.
We chose the aspects of our processing model so that we could study our hypothesis adequately.
Here, we motivate some aspects of the simplified naturalistic processing stream,
both with respect to biological findings, as well as to their functional role.

\emph{Vergence.}
As a first processing step, we incorporated vergence in our model,
the rotation of the two eyes towards each other.
The visual system controls gaze,
so that the image of objects or any structure of interest is moved to the fovea,
the location on the retina with the best spatial resolution.
Vergence is not used in common technical solutions.
State-of-the-art algorithms work with images obtained with parallel camera axes.
They calculate depth by applying epipolar stereo geometry
to corresponding locations in both half-images \cite{hartley_multiple_2004}.
However, if the goal is to understand vision based on statistical processing,
vergence has crucial impact.

Sparse coding is an optimization that builds on statistical regularities of the underlying data.
As a first approximation of stereo vision, each half-image is a locally shifted version of the other.
The extend of these image shifts, or disparities, differs broadly over the whole scene.
In contrast, if both eyes are oriented towards the same location,
the distribution of disparities in the vicinity of the fixation point is very narrowly distributed around zero,
as shown in Fig.~\ref{fig:DistributionOfDisparity_virtverg}.
This finding is due to the local smoothness of disparities,
disrupted only by discontinuities at object boundaries.
Only through vergence the sparse coding algorithm can find statistical dependencies between the two half-images,
because corresponding image locations are close-by.
Statistical dependencies then manifest in similarly shaped left and right half-kernels,
which are often slightly shifted versions of one another.
Indeed, \citet{hunt_sparse_2013} have shown
that sparse coding with simulated strabismus only extracts monocular dominant kernels.

\emph{Retinal pre-processing.}
Receptive fields of the retina are characterized by a center surround organization,
with weights in central location that are opposed in polarity to the weights around the center.
They are often modeled with the Mexican hat shaped Laplacian of a Gaussian,
or the simpler approximation with the difference of two Gaussians, like in our case.
Reasons discussed for this kind of retinal processing
include mechanisms of efficient coding, compression, response equalization, sparseness and others \cite{graham_can_2006}.
Convolving the visual input with a difference-of-Gaussians decorrelates the overall pink-noise spectrum of natural scenes,
transforming it into a representation with equalized power spectrum \cite{atick_what_1992}.
Removing these first order correlations, also called whitening,
is a common pre-processing step before applying an ICA procedure,
because it affects the algorithms' search for higher order statistical dependencies \cite{hyvarinen_independent_2000}.

\emph{Sparse coding in biological substrate.}
The sparse coding algorithm serves as a model for the formation of neuronal circuitry in V1.
It has been proposed that the gradient descent on the error function, with respect to the coefficients,
could be implemented directly in neural network topology
\cite{olshausen_sparse_1997,olshausen_principles_2003,rozell_sparse_2008}.
In the following, we consider single neurons for simplicity.
The gradient descent on coefficients \(a\), which is the activity of the neurons in neural network notion,
follows a differential equation.
In each time step,
the activity \(a\) of a neuron \(k\) changes proportionally to the sum of three terms
(see also Fig.~\ref{fig:ProcessingPipeline}):
(\emph{i.})\ a feed forward term \( \bm{\varphi}_{k}^\intercal \bm{x} \),
where the vectorized kernels \( \{ \bm{\varphi}_{k} \}_{k=1}^K  \) serve as receptive fields for the input \( {\bm{x}} \),
(\emph{ii.})\ a competition term \(- \sum_{c \neq k} \bm{\varphi}^{\intercal}_{k} \bm{\varphi}^{\vphantom{\intercal}}_{c} \, a^{\vphantom{\intercal}}_{c}\)
that introduces lateral inhibition proportional to the activity \(a_c\)
from all other neurons of the LCA layer,
with weights proportional to the similarity of the receptive fields,
and (\emph{iii.})\ a self-inhibition term \( -a_k \).
In LCA optimization, the sparse coding algorithm 
is extended by deriving a ``leaky integrator'' neuron (see \citet{abbott_lapicques_1999}).
The main difference is the introduction of an inner state \(u\),
which is coupled to the output of the neuron with a thresholding function \(a = T(u)\).
The three terms stay the same with LCA sparse coding, except that they drive the inner state \(u\) of the neuron
and that the self inhibition in term \emph{iii.}\
is replaced by \( -u \), the leak of the neuron.

This network is reminiscent
to the Hopfield network \cite{little_existence_1974,hopfield_neural_1982,hopfield_neurons_1984}.
With equivalent topology,
the weights of networks derived from the Hopfield network
are in many cases trained by applying biologically more plausible learning rules
that rely on information available at the synapse.
For example, F\"oldiak presented an artificial neural network
in which feed-forward weights were learned by Hebb's rule
and lateral inhibition was subject to anti-Hebbian learning.
Anti-Hebbian learning means
that inhibitory connections between neurons were enhanced if they were active at the same time \cite{foldiak_forming_1990}.
Therefore, the network learned competition between neurons that were driven by similar patterns,
akin to term \emph{ii}.
It was shown that a network with these learning rules, applied to natural images, developes Gabor-like kernels \cite{falconbridge_simple_2006}.
Applying the same Hebbian, anti-Hebbian learning to spiking neural networks yields similar results,
drawing even closer to a biologically accurate model \cite{zylberberg_sparse_2011,king_inhibitory_2013}.
\citet{chauhan_emergence_2018} applied such a network to stereo images
and reported successful disparity readout of the neural population with a simple classifier.
Physiological studies indeed provide evidence for lateral inhibition
between neurons with similar receptive fields in V1:
orientation selectivity of neurons
might benefit from lateral inhibition
between neurons with similar orientation tuning \cite{blakemore_lateral_1972}
or from other types of cross-orientation inhibition \cite{ringach_dynamics_2003,shapley_dynamics_2003}.

\emph{Probabilistic inference.}
Hypotheses on properties of the world are subject to uncertainty.
Bayesian inference provides a framework that allows to account for ambiguity
and a broad range of brain functions, like multimodal perception, decision making or motor control,
have been modeled following Bayesian approaches \cite{knill_bayesian_2004, doya_bayesian_2006}.
Training a perceptron-like neural network with backpropagation is linked to probabilistic inference.
With respect to stereo vision,
\citet{goncalves_what_2017} analyzed the relationship between a binocular likelihood model
and a two layer feed-forward neural network.
The first layer represented simple cells, preset with Gabor-like receptive fields,
and the second layer represented complex cells tuned for disparities.
The weights of both layers were trained with back-propagation.
The learned weights from simple to complex cells
werer proportional to the log-probability of the simple cell being active,
given the preferred disparity represented by the complex cell.
Because neural networks of this kind compute the weighed sum of the individual units' activities,
each complex cell calculated the log-likelihood of its preferred disparity.
This is equivalent to the na\"ive Bayes classifier we used for inferring disparity,
the logarithmic form of which can be implemented similarly in a neural network.

\section{Results}
Our analysis of the stereo-vision processing pipeline followed the main hypothesis of this paper:
that patterns from the external world
can be accessed with simple readout
from a representation obtained with sparse coding.
We chose disparity and surface orientation as candidates for such patterns.
In Sec.~\ref{sec:res:KernelCharacteristics},
we first describe qualitative and quantitative properties of the learned LCA representation.
Sec.~\ref{sec:res:disparity} addresses the main hypothesis of the paper
by evaluating the errors of simple readout of stereo disparity.
In Sec.~\ref{sec:res:overcompleteness} and \ref{sec:res:sparsity}
we discuss the extend of errors subject to overcompleteness and sparsity of the LCA representation.
The findings are expanded with Sec.~\ref{sec:res:predictingAccuracy},
where we describe how the accuracy of the inference can be predicted
by the overall activity in the LCA layer.
The mechanism holds implications for possible attention mechanisms.
Results from these subsections culminate
in the evaluation of disparity maps of naturalistic scenes in Sec.~\ref{sec:res:naturalistic}.
In Sec.~\ref{sec:res:orientatation} we then evaluate the orientation tuning of Kernels obtained by LCA optimization.

\begin{figure*}[b]
  \includegraphics[width=\textwidth]{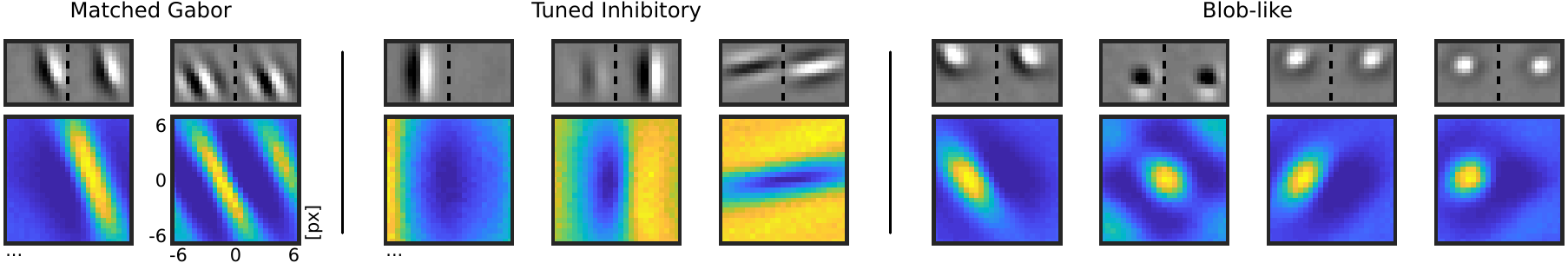}
  \caption{
    Typical kernels obtained by LCA optimization.
    \emph{Matched Gabor} left and right half-kernels were very similarly shaped,
    but shifted in position.
    \emph{Tuned Inhibitory} kernels were selective for large disparity values.
    They were mostly ocular dominant
    and left vs.\ right half-kernels were shifted in phase by about \(\pi\)~radians.
    \emph{Blob-like} kernels' weights consisted of a central spot
    and an outer lobe with opposed polarity.
    Disparity selectivity was more localized than for the other types.
  }
  \label{fig:ExampleKernels}
\end{figure*}

\begin{figure}
  \includegraphics[width=\columnwidth]{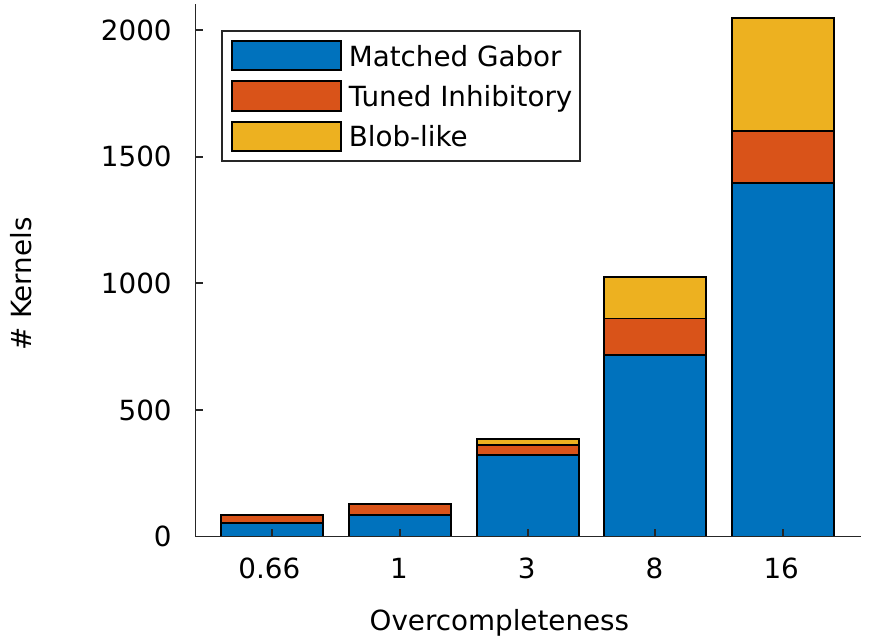}
  \caption{
    Proportion of the three kernel types 
    on the total number of kernels,
    plotted for each of the five trained values of overcompleteness.
    ``Matched Gabor'' and ``Tuned Inhibitory'' types 
    were evident on all levels of overcompleteness,
    with a decreasing fraction of ``Tuned Inhibitory'' types for larger models.
    ``Blob-like'' kernels only emerged in the models that were at least \(3 \times\)~overcomplete. 
  }
  \label{fig:KernelClassesQuantification}
\end{figure}

\subsection{Characteristics of the LCA representation} \label{sec:res:KernelCharacteristics}
In the following, we focus on results specific to disparity selectivity
and compare them to physiological findings.
As outlined in Sec.~\ref{sec:stereoICA}, the kernels
obtained by applying ICA methods to stereo image data
have been well described elsewhere.
We therefore limit our report to results specific to LCA sparse coding.

\subsubsection{Selectivity for disparity} \label{sec:res:KernelSelectivity}
For simple probabilistic readout,
individual neurons need to exhibit some degree of specificity for the pattern of interest.
Indeed, all tuning maps of kernels obtained with Eq.~\ref{eq:DisparityTuning}
yielded clear, smoothly varying selectivity as a function of disparity.
This was true throughout all kernels obtained by optimizing Eq.~\ref{eq:errorFct},
irrespective of the level of overcompleteness.
Therefore, all kernels potentially contribute to disparity inference with the simple readout scheme.
For representative examples, see Figs.~\ref{fig:ExampleInference}, \ref{fig:ExampleKernels} and \ref{fig:ComparisonPoggio}.
We include all learned kernels in the supplementary material, Figs.~S01--S05.
The shape of the kernels was in most cases well described by the Gabor function
(see Sec.~\ref{sec:res:statistics} and Fig.~\ref{fig:GaborFitAnalysisPanel}a,~c).
Kernels which were not Gabor-shaped,
and which were therefore not classical in terms of physiologically described receptive fields,
did only emerge with higher levels of overcompleteness.

We identified three main types of kernel shapes: ``Matched Gabor'', ``Tuned Inhibitory'' and ``Blob-like''.
A significant number of the ``Matched Gabor'' and the ``Tuned Inhibitory'' type 
were evident at all levels of overcompleteness.
However, the share of the ``Tuned Inhibitory'' type was decreasing
the larger the overcompleteness of the model.
The ``Blob-like'' type only emerged in models which were at least \(3 \times\)~overcomplete,
with increasing share the larger the overcompleteness of the model.
When presented with natural stereo images,
the average number of kernels from each type that contributed to the reconstruction of the stimulus
was proportional to their share in the set of kernels.
This finding was slightly violated if the sparsity penalty \(\lambda\) was very high.
In these cases, ``Matched Gabor'' kernels 
had an up to \num{10}\% larger share on the number of active kernels than on average.
The three types will be described in more detail in the following paragraphs.
For examples of each type see Fig.~\ref{fig:ExampleKernels}, 
for the share of each type on the total number of kernels see Fig.~\ref{fig:KernelClassesQuantification}.

\paragraph{Matched Gabor}
The majority of kernels were Gabor-like,
with very similar left and right shapes.
Differences between the two half-kernels
were best described by a shift in position and almost no shift in phase.
\citet{tsao_receptive_2003} reported that most receptive field shapes in V1
are also characterized by only a small amount of phase-shifts.
Such kernels are well suited
to represent corresponding (or matching) structures in the two half-images
that originate from the same object in the world (see also Sec.~\ref{sec:stereoBiology}).
Conversely, the mode of the tuning maps was equal to the position-shift of the two half-kernels.
Note that the mode was sharply peaked perpendicular to the orientation of the kernel shape,
but wide in direction of the orientation.
These kernels were therefore only selective for disparity perpendicular to their orientation.

\paragraph{Tuned Inhibitory}
The probability of these kernels being active
increased with the absolute value of disparity.
Typically, they were monocular or monocular dominant,
i.e., most of the weight energy was in either the left or the right half-kernel.
If they were binocular, 
the lobes were usually shifted by about \(\pi/2\) or by about \(\pi\)~radians (see also Sec.~\ref{sec:res:statistics}).
Such kernels were also reported by \citet{hunter_distribution_2015},
see Sec.~\ref{sec:stereoICA}.
Note that phase-shift kernels might serve as ``what not''-detectors when used for stereo inference,
as described in Sec.~\ref{sec:stereoBiology} \cite{read_sensors_2007,goncalves_what_2017}.
The weights of monocular kernels process information from only one stereo half-image.
An explanation for the disparity selectivity
based on feed forward processing is therefore unlikely.
With sparse optimization on the other hand,
matching structures can be reconstructed more sparsely with a single binocular kernel,
where otherwise two monocular kernels would be needed.
In neural network notion,
monocular and binocular kernels compete against each other through lateral inhibition.
The probability that a binocular kernel exists
that can jointly represent both half-images
decreases with larger disparities (see Fig.~\ref{fig:DistributionOfDisparity_LCA}).
Therefore, the likelihood that monocular kernels are active increases with disparity.

\paragraph{Blob-like}
The shapes of this type were not Gabor-like,
but had in common a center-surround organization
with a central spot of one polarity
and a surrounding structure with opposed polarity.
The surrounding lobe, however, varied in its extend,
not always completely enclosing the central spot.
The resulting shapes described a continuum,
with a partial opening resembling an end-stopped ridge,
an opening of approximately half extend matching a corner
and an even further opening describing slightly curved edges.
Kernels of this type were selective for disparity in both dimensions,
as opposed to Gabor-like kernels,
which were prone to the aperture problem:
the displacement of an oriented structure
can only be measured perpendicular to its orientation.
Our results reflect image statistics
and therefore show that natural images consist of a substantial amount of structures,
which are best described as corners, ridges and blobs.
Such elements may be used to reconstruct two-dimensionally displaced structures directly
rather than with a combination of local spatial frequency elements, i.e. Gabor-like kernels.
Note that \citet{ringach_spatial_2002} reported
a substantial amount of blob-like receptive field shapes in V1.

\begin{figure*}[t]
  \includegraphics[width=\textwidth]{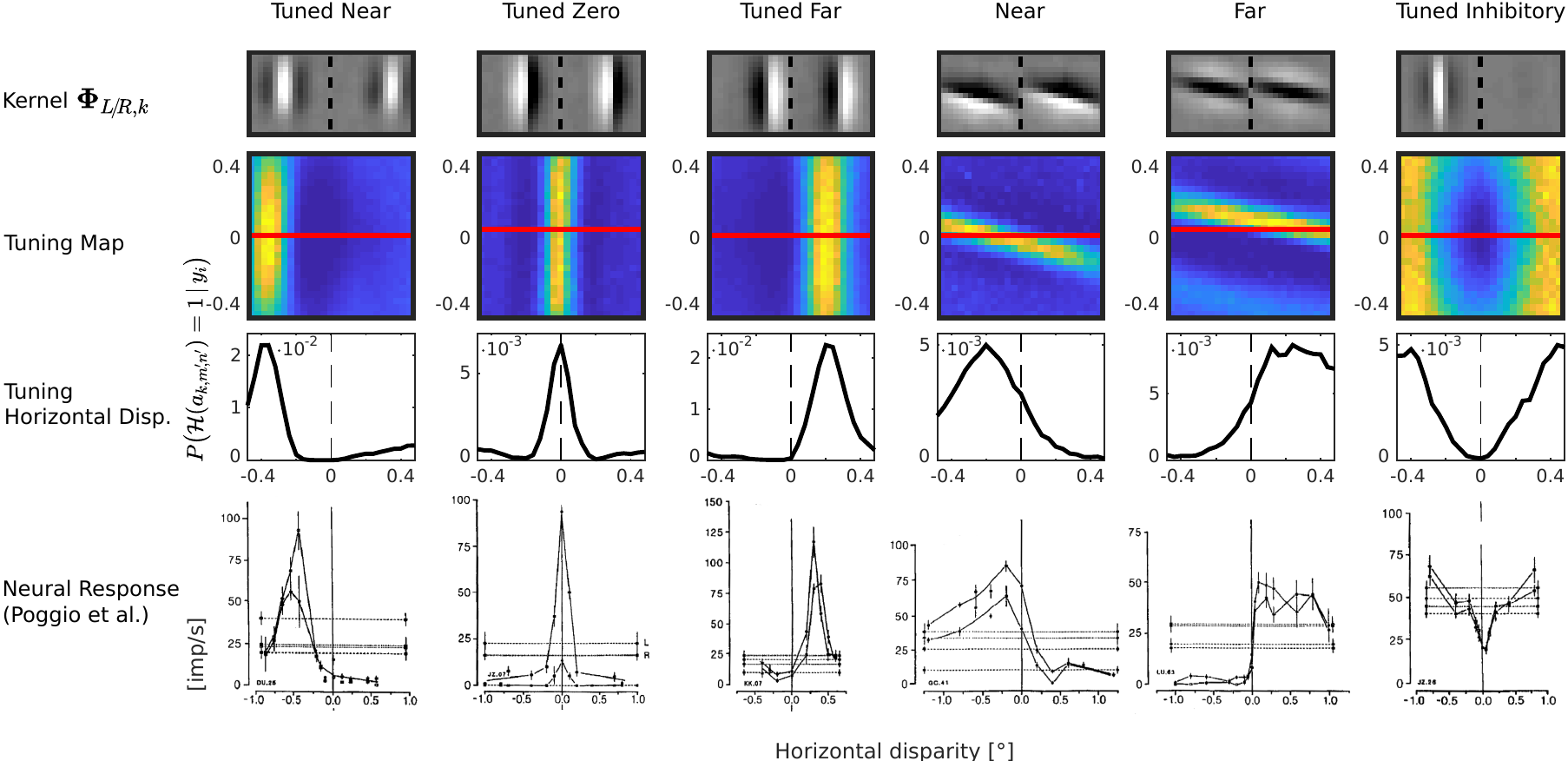}
  \caption{
    Comparison of example representatives from our data
    that match the six disparity response types defined by \citet{poggio_stereoscopic_1988}.
    Each column shows a single Kernel, its disparity tuning map
    and its horizontal cross section along the red line (horizontal disparity tuning).
    They fit the examples from the bottom row,
    which consists of the disparity tuning curves of physiological single neuron recordings
    from monkey visual cortex.
    All examples were drawn from the \(16 \times\)~overcomplete model, \(\lambda=0.04\).
    For details see Sec.~\ref{sec:res:KernelSelectivity}.
  }
  \label{fig:ComparisonPoggio}
\end{figure*}

The kernel shapes we obtained with LCA sparse coding fit well to physiological findings.
\citet{poggio_stereoscopic_1988}
recorded neuron responses from rhesus macaque monkey visual cortex
and classified disparity tuned cells in six categories.
Three of these, ``Tuned Near'', ``Tuned Zero'' and ``Tuned Far'' neurons,
were characterized by sharply peaked response curves,
tuned to negative, zero or positive horizontal disparities.
The two categories ``Near'' and ``Far'' contained neurons
that were similarly selective for negative or positive disparities.
However, these neurons' responses were not as peaked
as the responses of the ``Tuned'' neurons but rather broadly tuned.
The last category was referred to as ``Tuned inhibitory''
and contained neurons that were more likely to fire the larger the disparity,
irrespective of its sign.
We can reproduce the physiological examples of all six categories with our kernel sets
and present them in Fig.~\ref{fig:ComparisonPoggio}.

The three ``Tuned'' types describe the same response as our ``Matched Gabor'' kernels.
They were sharply tuned to disparity,
but only perpendicular to their orientation.
If oriented vertically,
they were therefore sharply tuned to horizontal disparity.
In some cases they were tuned to more than one disparity,
like in the second ``Matched Gabor'' example of Fig.~\ref{fig:ExampleKernels}.
This was due to the repetitions of the sinusoids.
However, most of our kernels had a single sinusoid lobe,
like in the ``Tuned Zero'' and ``Tuned Far'' examples of Fig.~\ref{fig:ComparisonPoggio},
which resulted in a single, elongated peak in the tuning maps.
It seems that single lobed kernels are a specialty of LCA sparse coding,
as compared to standard sparse coding.
This finding will be discussed in more detail in Sec.~\ref{sec:res:statistics}
(see also Fig.~\ref{fig:GaborFitAnalysisPanel}h).

We also found tuning maps which reproduce the ``Near''- and ``Far'' types.
Oblique ``Matched Gabor'' kernels
were more broadly tuned to horizontal disparity,
which was due to the kernels' elongated response peaks in two-dimensional disparity space.
However, we reproduced the horizontal tuning curves with shifted images,
which was a very stable stimulus.
We assume that horizontal disparity tuning of oblique kernels is very sensitive
to small changes of vertical disparity.
In addition, \citet{poggio_binocular_1977}
found that most ``Near''- and ``Far'' cells received unbalanced inputs from the two eyes,
which is not true for our ``Matched Gabor'' kernels.

The ``Tuned Inhibitory'' type matches our own classification.
In the respective paragraph we have offered an explanation
for how the lateral inhibition of the sparse optimization leads to tuned inhibitory units.
This finding has physiological support.
\citet{poggio_binocular_1977} and \citet{poggio_mechanisms_1981} reported that ``Tuned Inhibitory'' neurons
often showed ``strong excitatory dominance of one eye (ocular unbalance),
the `silent' eye exercising only inhibitory functions and only over a restricted disparity range''.
They also reported bidirectional cells,
``with balanced ocularity, from which stimulation of either eye alone evoked excitatory responses
that of the two eyes together evident response suppression''.
These bidirectional cell's responses were similar to the response of kernels
with about \(\pi\)~radians shifted sinusoid.
Further physiological evidence supports that suppressive mechanisms of this kind
help to solve the stereo correspondence problem \cite{tanabe_suppressive_2011,tanabe_delayed_2014,henriksen_disparity_2016}.
To the best of our knowledge,
tuned inhibitory units in stereo vision
have not been described in the context of sparse coding in the literature, yet.

\subsubsection{Statistical analyses of the kernels} \label{sec:res:statistics}

\begin{figure*}[t]
  \includegraphics[width=\textwidth]{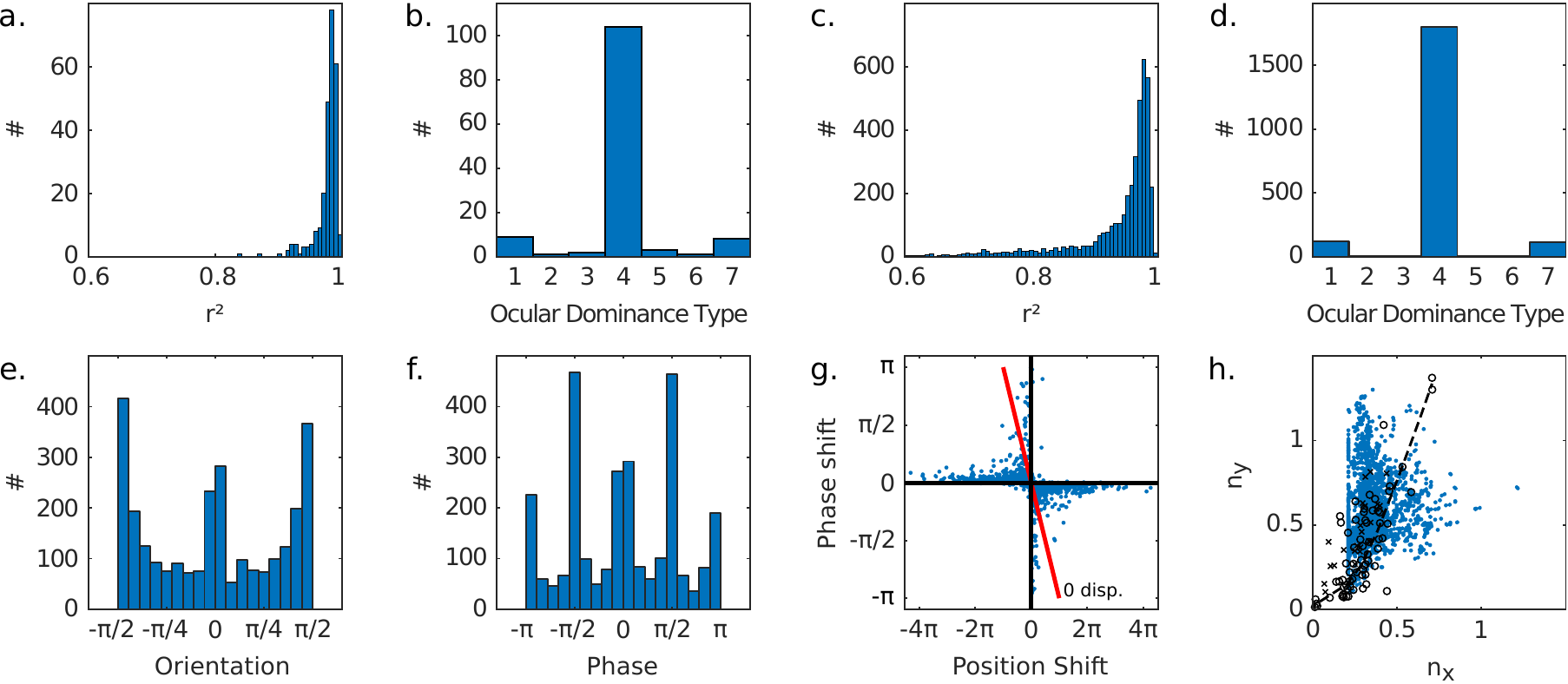}
  \caption{
  Statistics of the learned kernels~\({\bm{\Phi}_{\LR}}\),
  for details see Sec.~\ref{sec:res:statistics}.
  Diagrams {\bf a},~{\bf b}~contain data
  from the \(1 \times\)~overcomplete model (\num{128}~kernels),
  {\bf c}--{\bf h}~contain data
  from the \(16 \times\)~overcomplete model (\num{2048}~kernels).
    {\bf a},{\bf~c)}~Histogram of coefficients of determination as a measure for goodness of fit.
    {\bf b},{\bf~d)}~Ocular dominance types
  on \num{7} point scale:
  \num{4} is binocular; \num{1}~and~\num{7}~are left/right monocular, respectively.
  Row {\bf e}\--{\bf h}~is restricted to data with coefficient of determination \(r^2 > 0.93\),
  in order to reject non-classical receptive field shapes.
    {\bf e,~f)}~Distribution of orientation and phase.
    {\bf g)}~Interdependence between difference in position and phase of left vs.\ right Gabor fit.
  position-shift is expressed perpendicular to the orientation of the Gabor function
  and normalized by spacial frequency,
  calculated with \( f \, \lVert \left( \Delta x, \Delta y \right)^\intercal \rVert \cos{\phi}\).
  The red line marks zero-disparity.
  {\bf h)}~Relationship \(n_{\xy}=f \, \sigma_{\xy}\),
  with spatial frequency \(f\)
  and the width of the Gaussian envelope
  \(\sigma_{x}\)/\(\sigma_{y}\) (perpendicular/along orientation).
  Blue: our data.
  Black circles/crosses: data from macaque V1,
  reported by \citet{ringach_spatial_2002}/\citet{jones_evaluation_1987}.
  }
  \label{fig:GaborFitAnalysisPanel}
\end{figure*}

In order to characterize quantitative properties of the kernels from the LCA optimization,
we fitted the Gabor-function
\begin{equation}
  g(a,b,\phi,x,y,\theta,\sigma_x,\sigma_y)
  =
  a + b \,
  \exp\left(
  c
  \right)
  \cos\left(
  d
  \right)
\end{equation}
to each half-kernel \({\bm{\Phi}_{\LR,k}}\),
with offset \(a\), scale \(b\),
an elliptical Gaussian envelope
\( \exp\left(c\right) = \exp \left( \alpha x^{\prime \, 2} + 2 \beta x' y' + \gamma y^{\prime \, 2} \right) \)
and a sinusoid
\( \cos\left(d\right) = \cos \left( 2 \pi f x' + \kappa \right) \) along \(x'\),
with spatial frequency \(f\) and phase-shift \(\kappa\).
Orientation \(\phi\) and position \(x\),~\(y\)~in image space were free, with
\begin{equation}
  \begin{split}
    x' & = \hphantom{-} (x - x_0) \, cos(\phi) \,+\, (y - y_0) \, sin(\phi) \, , \\
    y' & =           -  (x - x_0) \, sin(\phi) \,+\, (y - y_0) \, cos(\phi) \, .
  \end{split}
\end{equation}
The elliptical envelope, with widths \(\sigma_x\) and \(\sigma_y\), was allowed to rotate freely
by the angle \(\theta\),
relative to the orientation of the sinusoid,
with
\begin{equation}
  \begin{split}
    \alpha & = \hphantom{-} \frac{cos(\theta)^2}{2 \sigma_x^2} + \frac{sin(\theta)^2}{2 \sigma_y^2} \; , \\
    \beta  & =           -  \frac{sin(2 \theta)}{4 \sigma_x^2} + \frac{sin(2 \theta)}{4 \sigma_y^2} \; , \\
    \gamma & = \hphantom{-} \frac{sin(\theta)^2}{2 \sigma_x^2} + \frac{cos(\theta)^2}{2 \sigma_y^2} \; .
  \end{split}
\end{equation}
We used a custom implementation in MATLAB,
which we made publicly available\footnote{\url{https://www.mathworks.com/matlabcentral/fileexchange/60700-fit2dgabor-data-options}}.

Most kernels were well described by the Gabor function,
with the coefficient of determination \(r^2\) close to \num{1} (Fig.~\ref{fig:GaborFitAnalysisPanel}a,~c).
Some of the lower values can be attributed to monocular kernels,
in which the half-kernel with less weight energy has a lower signal-to-noise ratio.
With higher levels of overcompleteness, more non-Gabor-like kernel shapes appeared,
which is apparent with the heavy tail in the distribution
of the \(16 \times\)~overcomplete model (\num{2048}~kernels) in Fig.~\ref{fig:GaborFitAnalysisPanel}c,
as opposed to the distribution
of the \(1 \times\)~overcomplete model (\num{128}~kernels) in Fig.~\ref{fig:GaborFitAnalysisPanel}a.

We analyzed the ocular dominance of the kernels
by adapting the \num{7}~point scale from Hubel and Wiesel \cite{hubel_receptive_1962}.
They were calculated with
\begin{equation}
  \arctan \left( \frac{\lVert \bm{\Phi}_{{L,k}} \lVert} {\rVert \bm{\Phi}_{{R,k}} \rVert} \right) \, ,
\end{equation}
with values in the range \([0,\pi / 2]\) plotted in a histogram with \num{7} equally spaced bins.
Kernels which were left or right monocular fell into category \num{1}~and~\num{7}, respectively.
If weight energy was equally distributed, kernels fell into category \num{4},
the other categories were left or right dominant, respectively.
We show the results in Fig.~\ref{fig:GaborFitAnalysisPanel}b and~d.
The majority of kernels were in category \num{4},
i.e., binocular with balanced weight energy (see Fig.~\ref{fig:GaborFitAnalysisPanel}b,~d).
A substantial fraction of kernels was purely monocular (category \num{1}~or \num{7}).
Only a small fraction was in the intermediate categories
and the proportion of intermediate kernels was even lower with higher levels of overcompleteness.
The shape of these kernels was usually characterized by a phase-shift of about \(\pi\)~radians.
Kernels that did not fall into category \num{4}~were usually of the ``Tuned Inhibitory'' type,
described in Sec.~\ref{sec:res:KernelSelectivity}.
In physiological experiments,
a similar three-mode distribution of weight energy between left and right receptive fields
was also found in ferrets,
albeit not as distinctly peaked as in our results. \cite{kalberlah_sensitivity_2009}.
Other physiological studies on various animals report rather flat distributions \cite{schiller_quantitative_1976,hubel_receptive_1962,levay_ocular_1978,guillemot_binocular_1993,hubel_binocular_2015}.

The following analyses were based on the \(16 \times\)~overcomplete model.
Fits with a coefficient of determination of \(r^2 < 0.93\) were excluded
in order to exclude non-classical receptive field shapes.
The distribution of orientations (Fig.~\ref{fig:GaborFitAnalysisPanel}e)
had two peaks at \(0\)~degrees and at \(\pm 90\)~degrees.
Two possible explanations have been offered in the literature for this bias:
the rasterization of the input images and the prevalence of orientations
in human made structures \cite{hunt_sparse_2013}.
The distribution of phases (Fig.~\ref{fig:GaborFitAnalysisPanel}f)
had distinct peeks at \(0\)~degrees, at \(\pm 90\)~degrees and at \(\pm 180\)~degrees,
i.e., the kernel shapes were, in most cases, either sine-like or cosine-like.
\citet{ringach_spatial_2002} reported that physiological receptive fields
similarly cluster in such even- and odd-symmetric shapes.
Opposed to our findings with LCA sparse coding,
he also reported that, with standard sparse coding,
there is a tendency towards odd-symmetric receptive fields,
but not towards even-symmetric receptive fields.

As described in Sec.~\ref{sec:stereoBiology},
binocular Gabor-filters that are shifted in position from left to right half-kernel
can serve as matched filters for corresponding image structures,
whereas Gabor-filters shifted in phase
can serve as ``what not''-detectors for false matches \cite{read_sensors_2007,goncalves_what_2017}.
We were therefore interested in the interrelationship
between the shift in position and the shift in phase of the kernels in our data.
Results are displayed in Fig.\ref{fig:GaborFitAnalysisPanel}g.
Because the tuning maps were characterized by elongated peaks,
we expressed the position-shift relative to the most sharply tuned axis.
It was therefore calculated
as the difference in horizontal and vertical position,
projected on the axis perpendicular to the orientation of the Gabor function.
For better comparability between position-shift and phase-shift,
we also normalized the position-shift by the spatial frequency of the sinusoid.
The position-shift was therefore calculated as
\( f \, \lVert \left( \Delta x, \Delta y \right)^\intercal \rVert \cos{\phi}\).
Our data
showed a transient separation between position-shift and phase-shift kernels.
If a kernel had both, a substantial position- and phase-shift, they counteracted each other,
so that almost all data points fell into quadrant {\romannumeral 2} and {\romannumeral 4}.
Lobes of the sinusoid match when data points are on the red line.
The majority of the kernels was mainly shifted in position
and therefore match the ``Matched Gabor'' type from Sec.~\ref{sec:res:KernelSelectivity}.

\citet{ringach_spatial_2002} reported that Gabor-like receptive field shapes of macaque V1
were more variable and often more blob-like
than kernels from sparse coding and basis vectors from ICA.
In his study,
he related the spatial frequency \(f\) of the sinusoid
to the extend of the Gaussian envelope
\(\sigma_{x}\),~perpendicular to the orientation of the sinusoid,
and \(\sigma_{y}\),~along the orientation of the sinusoid.
The relationship \(n_{\xy} = f \, \sigma_{\xy}\)
was lower on average in physiologically measured receptive fields.
In Fig.~\ref{fig:GaborFitAnalysisPanel}h,
we show an overlay of the data adapted from \citet{ringach_spatial_2002} (macaque V1, black circles),
and from \citet{jones_evaluation_1987} (cat V1, black crosses),
with our data (blue dots).
In this case, we fit the Gabor-functions
with the orientation of the elliptical Gaussian envelope fixed at \(\theta = 0\)~degrees.
While standard sparse coding and ICA results in values \(n_{\xy} > 0.5\)
for the majority of kernels / basis vectors,
many kernels from convolutional LCA sparse coding were characterized by lower values.
Note that we bound the fitting procedure to \(n_{\xy} \geq 0.25\)
and did therefore not allow blob-like fits,
so that these kernels do not appear in the panel.
The plot also shows that physiological receptive fields,
as well as the LCA kernels,
had a tendency for \(n_y > n_x\),
which was not true for standard sparse coding and ICA, as reported by Ringach.
Kernels with small values for \(n_x\),
i.e., with a small extend of the Gaussian envelope perpendicular to their orientation,
are better suited for disparity inference.
If the kernel shape consisted of only one sinusoidal lobe (\(n_x = 0.25\)),
the associated tuning map had a single elongated peak,
as opposed to kernels with more than one sinusoidal lobe,
which had multiple, parallel, elongated peaks.
The disparity they represented was therefore not ambiguous.
Indeed, we observed aliasing effects in the disparity inference
if image structures were represented by multi-lobe kernels.
For examples, see both ``Matched Gabor''-kernels from Fig.~\ref{fig:ExampleKernels}.

\subsection{Evaluation of disparity inference} \label{sec:res:disparity}

\begin{figure}
  \includegraphics[width=\columnwidth]{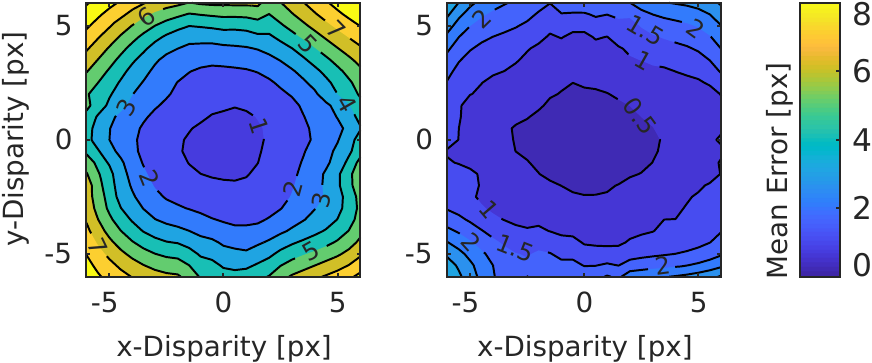}
  \caption{
    Mean error of absolute disparity estimates
    as a function of x- and y-disparity.
    \emph{Left}: \(1 \times\)~overcomplete, \(\lambda=0.04\).
    \emph{Right}: \(16 \times\)~overcomplete, \(\lambda=0.04\).
    Disparity estimates were evaluated
    with stereo images, in which the left half-image was a shifted version of the right half-image.
    With more overcompleteness in the LCA representation,
    the error for large disparities decreased.
  }
  \label{fig:ErrorOfDisp}
\end{figure}

\begin{figure}
  \includegraphics[width=\columnwidth]{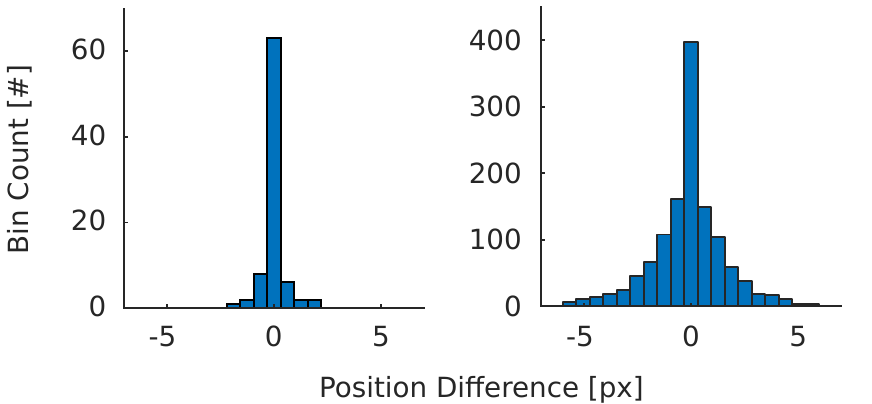}
  \caption{
    Distribution of the difference in position of ``Matched Gabor'' kernels.
    position-shift is expressed perpendicular to the orientation of the Gabor function.
    The plot includes all kernels with \(r^2 > 0.93\) and \(\phi < 0.3\)~rad.
    \emph{Left}: \(1 \times\)~overcomplete, \num{84} of \num{128} kernels.
    The kurtosis of the distribution is \(k = 9.26\).
    \emph{Right}: \(16 \times\)~overcomplete, \num{1264} of 2048.
    The kurtosis is \(k = 5.12\).
    LCA optimization with more overcompleteness
    yields kernels that represent a wider range of disparities.
  }
  \label{fig:DistributionOfDisparity_LCA}
\end{figure}

\begin{figure*}[t]
  \includegraphics[width=\textwidth]{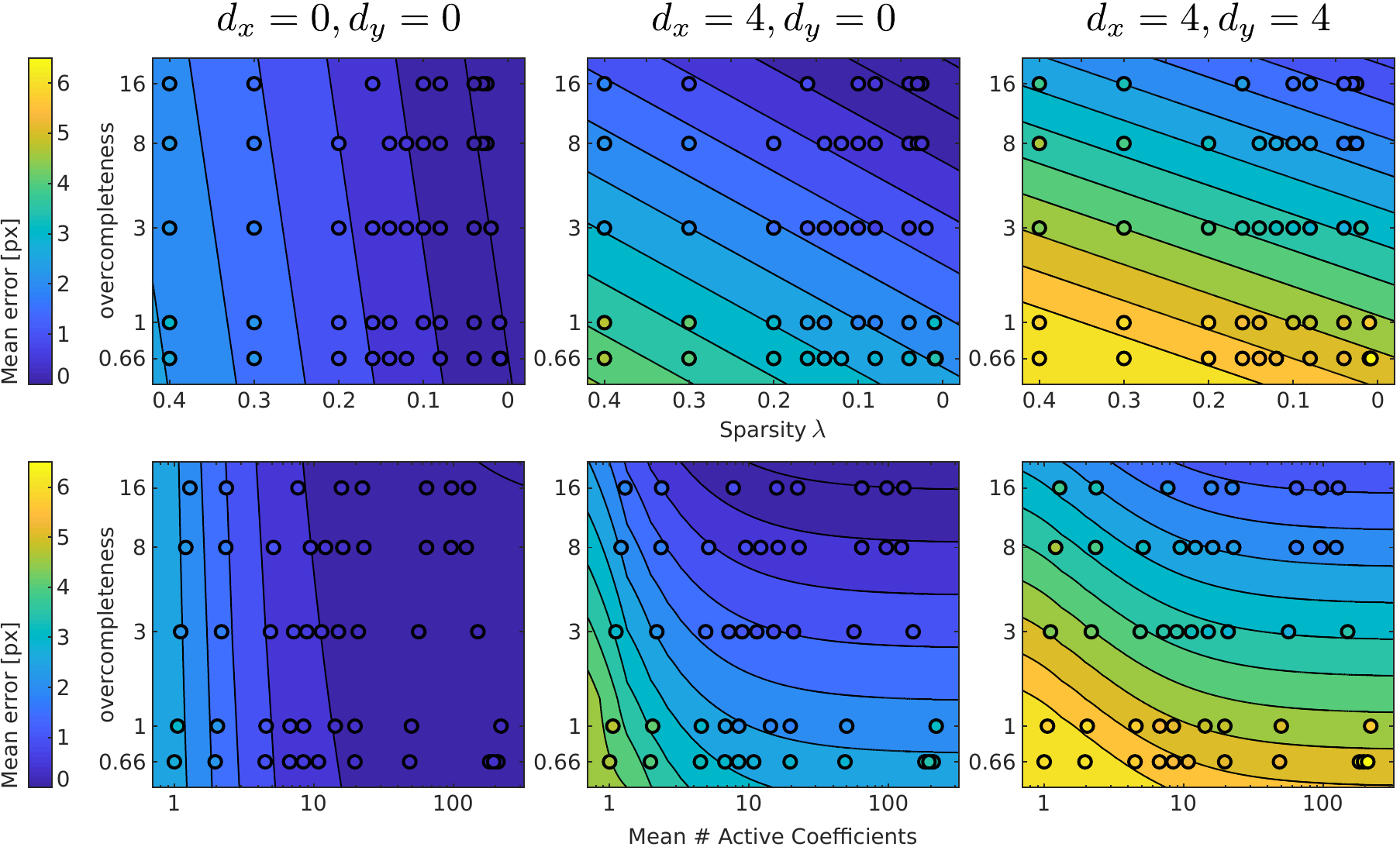}
  \caption{
    Top row panels show dependency of the mean inference error of disparity
    on overcompleteness (ordinate) and on sparsity load \(\lambda\) (abscissa).
    Bottom row panels show the same data
    but with \(\lambda\) mapped to the mean number of active coefficients.
    The three columns contain evaluations for three different disparities \(d_{x,y}\).
    See Fig.~\ref{fig:ErrorOfDisp}
    for the dependency of the mean error on disparity.
    Circles indicate evaluations of the mean absolute error (MAE)
    subject to overcompleteness \(o\) and  sparsity load \(\lambda\).
    Contours show the data fits of the error,
    with \(\text{MAE} = a\,(\lambda+\Delta_\lambda)\,+b\,(\ln{o}+\Delta_o)\,+\,c\,\) (top row)
    and \(\text{MAE} = a/({n+\Delta_n})\,+b\,(\ln{o}+\Delta_o)\,+\,c\,\) (bottom row).
    Close to zero disparity, overcompleteness has little impact,
    but it becomes increasingly important for larger disparities.
    The error generally declines with decreasing lambda.
    See the same data as line plots in supplementary Figs.~S01--S06.
  }
  \label{fig:MeanError_SparsityAndOvercompleteness}
\end{figure*}

In this subsection we evaluate
whether disparities can successfully be obtained with simple readout from  the LCA representation.
We explored the limitations by means of the error of the estimates.
Inference of disparity was carried out with the full processing pipeline,
subject to overcompleteness and sparsity penalty in the LCA optimization as described in Sec.~\ref{sec:LCA},
and with probabilistic readout as described in Sec.~\ref{sec:inference}.
The mean absolute errors (MAE) of the estimates were calculated with
\begin{equation}
  \text{MAE} = \frac{1}{n} \sum_{j=1}^n \, \lVert \, {\bm y}_j - \hat{\bm y}_j \rVert \, ,
\end{equation}
where \({\bm y}_j\) and \(\hat{\bm y}_j\) were the ground truth and the the estimate, respectively.
In Sec.~\ref{sec:res:overcompleteness}--\ref{sec:res:predictingAccuracy},
we report the MAE of the disparity estimates
\( \hat{\bm y}_j = {( \hat{d}_x \,\, \hat{d}_y )}^\intercal \).
Inference was carried out on  the test set from the disparity database with shifted images,
described in detail in Sec.~\ref{sec:stereoDatabases}.
In Sec.~\ref{sec:res:naturalistic} we report results on inference of horizontal disparity
in naturalistic stereo images.

\subsubsection{Higher dimensionality extends the set of detectable patterns} \label{sec:res:overcompleteness}

An increase of overcompleteness
generally resulted in a decrease of disparity inference errors.
The best parameter combination from our evaluation
(\(16 \times\)~overcomplete, \(\lambda=0.04\))
allowed for a mean disparity error below \num{0.5}\,px,
measured within the range of \(\sim\)\num{2}--\num{3}\,px ground truth absolute disparity (Fig.~\ref{fig:ErrorOfDisp}, right panel).
Inference was better for small disparities than for large disparities.
The same model performed with an error of \(\sim\)\num{1.5}\,px
for disparity of \(d_x = 4\)\,px horizontally
and \(d_y = 4\)\,px vertically.
The bias was generally small (data not shown)
and apparent only at large disparities close to the cut-off at \num{6}\,px.

Overcompleteness had its main impact on
the range of disparities for which the model performed well.
With small overcompleteness,
the error increased much more rapidly with the value of disparity.
For example, the \(1 \times\)~overcomplete model with \(\lambda=0.04\)
evaluated with an error below \num{1}\,px
within the range of \num{+- 1}\,px disparity
but with an error \(\sim\)\num{6}\,px
at \(d_{x,y}=4\)\,px horizontal and vertical disparity (Fig.~\ref{fig:ErrorOfDisp}, left panel).
We show all parameter combinations we tested
in the overview in Fig.~\ref{fig:MeanError_SparsityAndOvercompleteness}.
The same data is shown as line plots in supplementary Figs.~S01--S06.
For low levels of overcompleteness (\(d_{x,y}=0\)\,px, left-hand column of the plot),
error dependency on overcompleteness is negligible,
whereas  for large overcompleteness (\(d_{x,y}=4\)\,px, right-hand column),
overcompleteness has a substantial effect.

This finding was due to qualitative differences in the sets of learned kernels.
With larger overcompleteness,
more kernels existed with larger position-shift,
i.e., with the differences in position between left and right half-kernel (Fig.~\ref{fig:DistributionOfDisparity_LCA}).
The distribution of the kernels' disparity was roughly similar
to the distribution of disparities in stereo images (see Fig.~\ref{fig:DistributionOfDisparity_virtverg}),
with many kernels that represent small disparities and few kernels that represent large disparities.
We conclude that, with more overcompleteness,
sparse coding extends the set of patterns that are represented explicitly,
ordered by the frequency of their occurrence.

\subsubsection{Less sparsity results in lower errors} \label{sec:res:sparsity}

\begin{figure*}[t]
  \includegraphics[width=\textwidth]{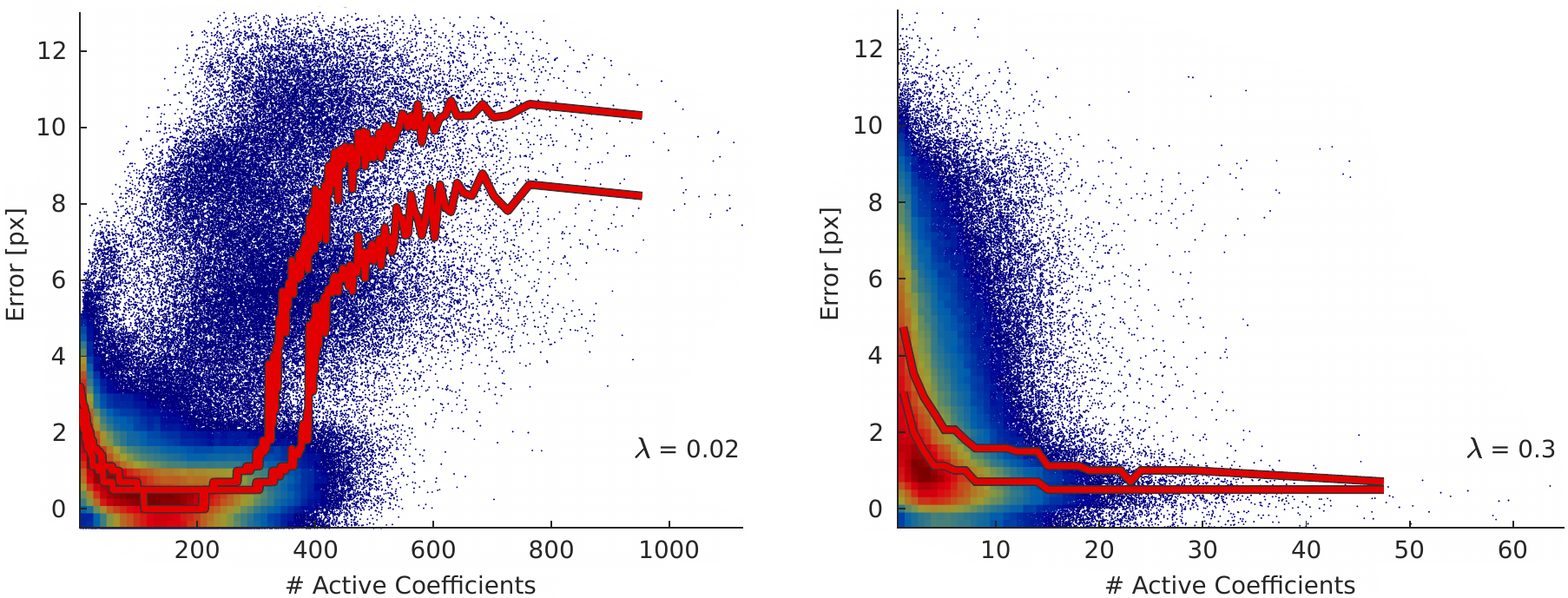}
  \caption{
    Data from \(3 \times\)~overcomplete model.
    The error of the inference
    is related to the mean number of active LCA coefficients.
    Data points represent the errors of single disparity inferences
    against the number of active coefficients.
    To counteract rasterization,
    they were displaced randomly by a small amount.
    The heat-map overlay is a density histogram (arbitrary units).
    Red lines: median and \(75\)th percentile of error,
    calculated on bins of the number of active coefficients (at least \(10^3\) data points per bin).
    \emph{Left}: With low sparsity penalty \(\lambda=0.02\),
    mid-range activity predicts the lowest error,
    as opposed to a small, or a large number of active coefficients.
    \emph{Right}: With increased sparsity load \(\lambda=0.3\),
    a larger number of active neurons is no longer associated with poor performance.
    Note that overall activity is substantially reduced.
  }
  \label{fig:ErrorDepOnN_detail_o3}
\end{figure*}

The sparsity load \(\lambda\) was generally linked to better inference the \emph{lower} its value.
Up to a limit of very low values for \(\lambda\),
this is true for all levels of overcompleteness and for all ground truth values of disparity,
as can be seen in the top row of Fig.~\ref{fig:MeanError_SparsityAndOvercompleteness}.
Our results are in line with the results from \citet{rigamonti_are_2011}
and from \citet*{gardner-medwin_limits_2001} (see Sec.~\ref{sec:related}).
The bottom row of Fig.~\ref{fig:MeanError_SparsityAndOvercompleteness}
contains the same data as the top row, but with the sparsity load \(\lambda\)
mapped to the mean number of active coefficients.
Activity was roughly linked by a negative exponential to the range of \(\lambda\) we tested.

In all models except the \(16 \times\)~overcomplete model,
we observed slightly increasing errors
if sparsity load was very low.
For most combinations of overcompleteness and disparity that we evaluated,
the lowest mean error was measured at \(\lambda \approx 0.04\).
The error was below \(\lambda = 0.1\) in all cases but one---%
the \(0.66 \times\)~overcomplete model,
measured at \num{4}\,px horizontal disparity and \num{0}\,px vertical disparity.
The minima can be examined in detail in the supplementary Figs.~S01--S06.
We therefore reject the hypothesis that 
inference is optimal if the sparsity penalty used during testing matches that used during training.
A possible explanation is based on the fact
that a binary multi channel code carries most information
if the probability of the coefficients to be in one of both states is \(p=0.5\)
and independent of other dimensions.
\cite{shannon_mathematical_1948}.
Therefore, assuming that the coefficients were independent,
the code carried most information
if half of the coefficients were in an active state on average
(\num{42.5} for \(0.6 \times\)~overcomplete,
\num{64} for \(1 \times\)~overcomplete,
and \num{192} for \(3 \times\)~overcomplete).
Inference was best slightly below these numbers,
which shows in the bottom row of Fig.~\ref{fig:MeanError_SparsityAndOvercompleteness}.
It was not possible to confirm this finding for larger overcompleteness
or larger values of \(\lambda\),
because very high sparsity load was computationally prohibitive.

The reasoning that Shannon information is the limiting factor is ambivalent.
On the one hand,
imposing less weight on sparsity in Eq.~\ref{eq:errorFct}
in turn imposes more weight on the reconstruction constraint,
and therefore the preservation of information.
On the other hand,
information of an overcomplete representation is highly redundant.
It is opposed to a compressed code that maximizes Shannon entropy \cite{field_what_1994}.
However, we binarized the output of the LCA sparse coding before inference,
which removed much information from each dimension.
Therefore, information content was strongly limited
if only a few coefficients were in an active state.

\begin{figure*}[t]
  \includegraphics[width=\textwidth]{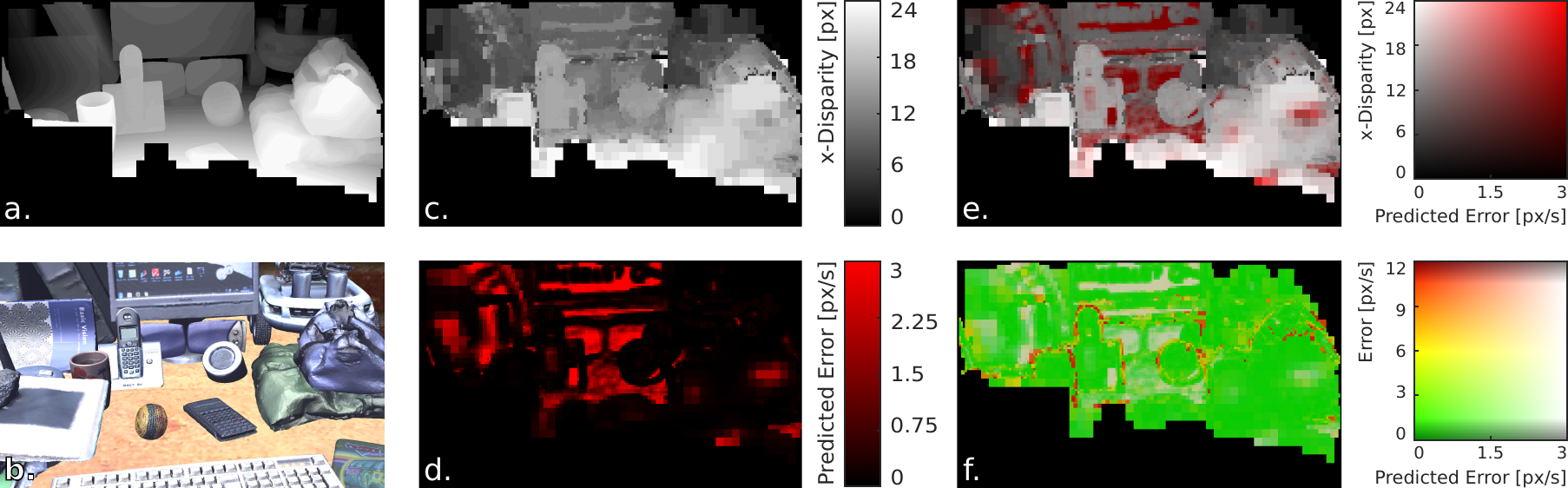}
  \caption{
  Disparity inference of a naturalistic scene
  from the Genua Pesto database \cite{canessa_dataset_2017}.
  {\bf a)}~Ground truth map of horizontal disparity.
  Values outside the range \([-24,24]\)\,px were excluded.
    {\bf b)}~Right stereo half-image.
  {\bf c)}~Inference of horizontal disparity
  with the \(16 \times\)~overcomplete, \(\lambda=0.04\) processing pipeline.
  Due to the scale space approach, resolution is better the closer objects are to the horopter.
    {\bf d)}~Errors predicted by the number of active coefficients \(a_{k,m'\!\!,n'}\),
  relative to scale \(s=1\), \num{.75}, \num{.5}, \num{.25} in subplots {\bf d}--{\bf f}.
  {\bf e)}~Overlay of disparity map {\bf c} and predicted error {\bf d}.
  {\bf f)}~Error of inference vs.\ predicted error.
  Both values are strongly correlated
  with \(r=0.58\) for predicted errors \(> 1\)\,px.
  }
  \label{fig:GenuaResults}
\end{figure*}

\subsubsection{The number of active LCA coefficients predicts the accuracy of inference} \label{sec:res:predictingAccuracy}
We encountered a strong relation between
the success of disparity inference
and the number of active coefficients.
We assessed this relation
by sorting responses to examples from the disparity database test set,
ordered by the number of active LCA coefficients \(a_{k,m'\!\!,n'}\).
The data were binned with a window size of at least \(10^3\) data points.
Note that the bin size was unequal, due to this constraint.
Finally, we calculated percentiles of the MAE of disparity inference.
Resulting histograms
are shown as the red lines in Fig.~\ref{fig:ErrorDepOnN_detail_o3}.
The data points of the disparity inference error are plotted in the same diagram,
with a heatmap overlay that displays density where the point cloud is very dense (arbitrary units).

The median error as a function of the number of active coefficients was u-shaped.
Therefore, inference was best when an average number of coefficients was in an active state.
A low number, as well as a large number of active coefficients was a predictor for large errors
(Fig.~\ref{fig:ErrorDepOnN_detail_o3}, left panel).
With a large value of sparsity load \(\lambda\),
the number of active coefficients was greatly reduced (right panel).
In this case, the median error was monotonically decreasing
as a function of the number of active coefficients.

We assume that a low number of active coefficients was associated to large errors
because the few tuning maps did not contain enough information for accurate inference.
The finding could simply account for the absence of structure in the image.
An explanation for the association of a large number of active coefficients with large errors
is not so straight forward.
We hypothesize that the sparse optimization
was not able to settle on a good representation
and therefore reconstructed the input with much more kernels than on average.
These kernels were not well suited for the given image structures
and therefore only active due to the lack of better representatives.
Our perspective is linked to a study from \citet{froudarakis_population_2014}.
They report that the stimulation with phase scrambled movies
activates mouse V1 more strongly than the stimulation with natural movies.
Simultaneous recordings from a large population of cells
were analyzed for discriminability of the presented movies
with a linear classifier.
Similar to our finding, strong activation was a predictor for bad classification performance.

\subsubsection{Disparity map of a naturalistic scene} \label{sec:res:naturalistic}

In addition to inference with constant disparity,
i.e., with shifted images,
we evaluated our visual processing pipeline with a naturalistic scene
from the Genua Pesto database \cite{canessa_dataset_2017}.
We present results from one of the scenes in Fig.~\ref{fig:GenuaResults}.
It consists of disparities in the interval \([-76.7,77.1]\)\,px,
as opposed to our model, which is limited to inference in the interval \([-6,6]\)\,px.
We faced the limitation of the model with a scale-space approach,
by downsampling the input image to \num{80~}\%, \num{60~}\%, \num{40~}\% and \num{20~}\%.
Inference was then only evaluated within the interval \([-6,6]\)\,px at each of these four scales,
and with the best available spatial resolution at each location.
Image locations outside the interval were excluded beforehand.
All experiments were carried out with the \(16 \times\)~overcomplete, \(\lambda=0.04\) model.

Disparity was inferred well within the aforementioned limitations.
The disparity map we obtained is shown in Fig.~\ref{fig:GenuaResults}c
(compare to ground truth disparity map in Fig.~\ref{fig:GenuaResults}a).
Note that the map is an overlay of the four scales,
with best spatial resolution close to the horopter.
We chose to plot the predicted error (Fig.~\ref{fig:GenuaResults}d)
as the \num{80~}th percentile of the error as a function of the number of active coefficients,
as described in Sec.~\ref{sec:res:predictingAccuracy}.
Errors in subplots d--f are relative to the scale \(s=1\), \num{.75}, \num{.5}, and \num{.25}.
Fig.~\ref{fig:GenuaResults}e is an overlay of the disparity map with the predicted error.
The prediction corresponds to clearly identifiable structures in the image.
Large errors were predicted for the loudspeakers, for the table texture,
and for uniformly colored locations on the monitor.
Low errors were predicted for the telephone, for the bags on the right,
and for the icons on the monitor.
Fig.~\ref{fig:GenuaResults}f visualizes the predicted error
and the actual error with respect to ground truth disparity in one plot.
If prediction failed, this was mostly due to occlusion boundaries.
Note that occlusion boundaries were not part of the training,
so this type of error can not be attributed to the lack of representation in the LCA optimization.

\subsection{Tuning maps of surface orientation} \label{sec:res:orientatation}

\begin{figure*}[t]
  \includegraphics[width=\textwidth]{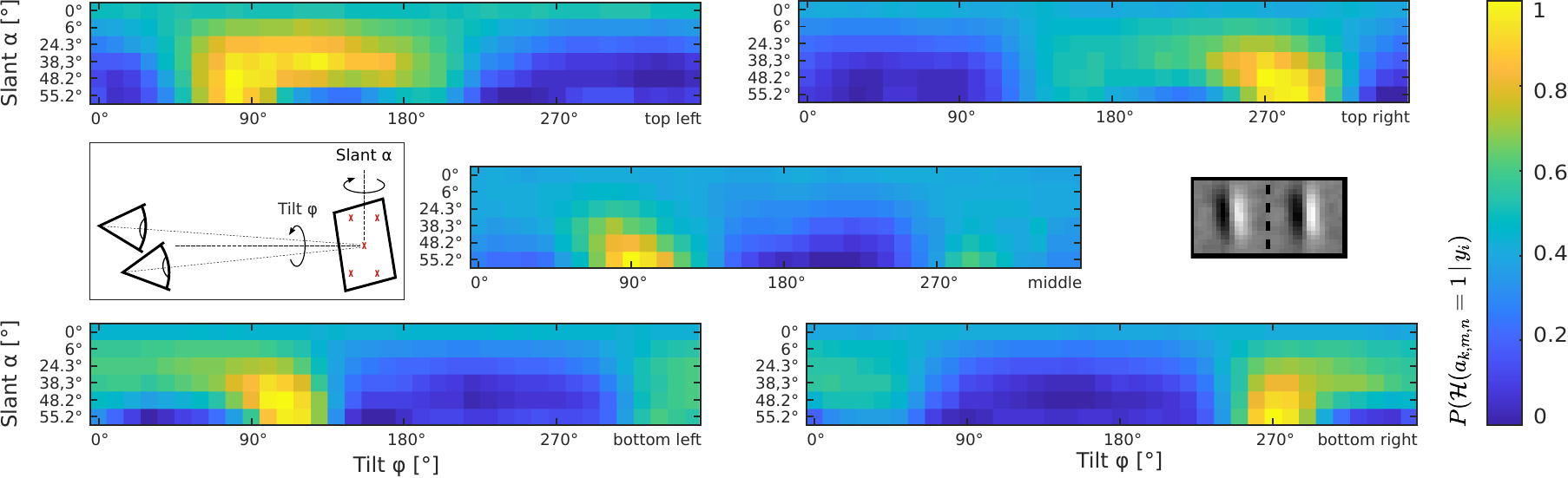}
  \caption{
    Tuning maps for tilt- and slant angles \(\varphi\) and \(\alpha\) of textured surfaces.
    All maps correspond to a single kernel \( \bm{\Phi}_{\LR,k} \) (grayscale)
    but with receptive fields
    at varying feature map locations
    w.r.t.\ the central fixation point.
    Feature map locations of the four tuning maps
    are indicated with red crosses in the illustration.
    Data from \(1 \times\)~overcomplete model, with \(\lambda=0.12\).
  }
  \label{fig:SlantTiltTuningMaps}
\end{figure*}

We showed that a representation formed by LCA sparse coding
forms a suitable basis to infer stereo disparity.
However, we hypothesized that sparse coding fulfills the requirement for simple readout
of a much larger set of patterns.
As a second example, we examined tuning maps
for tilt- and slant angles \(\varphi\) and \(\alpha\) of a textured surface
(see Sec.~\ref{sec:stereoBiology}).
Results were based on the test set from the surface orientation database,
described in detail in Sec.~\ref{sec:stereoDatabases}.

We created tuning maps, not only for each kernel \( \bm{\Phi}_{\LR,k} \),
but for each of \num{7x7} entries from the convolutional feature maps with central fixation point.
This decision was based on the expectation
that the tuning maps were affected
by the disparity tuning of the kernels.
We reasoned that surface orientation could be inferred
from a set of disparity measurements at positions relative to the fixation point.
In Fig.~\ref{fig:SlantTiltTuningMaps} we show
that kernels were tuned for surface tilt- and slant angles.
Indeed, the tuning maps of the kernels differed,
depending on the position at which it was evaluated.
The peak of the tuning maps was the sharper the larger the slant angle of the surface.
In tilt-/slant space, the peak described a skewed band,
which was expected if disparity tuning was the underlying principle.

\begin{figure}
  \includegraphics[width=\columnwidth]{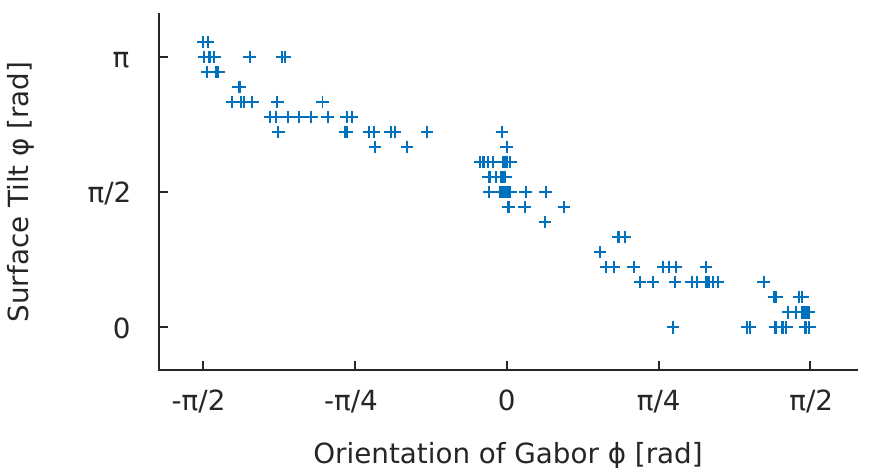}
  \caption{
    The tuning maps' mode of the tilt angle \(\varphi\)
    against the orientation \(\phi\)
    of kernels \( \bm{\Phi}_{\LR,k} \).
    All tuning maps were evaluated at the central fixation point
    with zero disparity.
    Orientation of the Gabor-like kernels
    accounts for surface tilt tuning,
    with very strong circular-circular correlation \(\rho_{cc}=-0.981\).
  }
  \label{fig:GaborRotVsTiltSelectivity}
\end{figure}

Coefficients in the center of the tuning maps,
which corresponded to the fixation point,
were also clearly tuned for the surface tilt angle
(see central tuning map of Fig.~\ref{fig:SlantTiltTuningMaps} for an example).
They could not be affected by disparity
because disparity is zero at the fixation point,
irrespective of the surface orientation.
Instead, the mode of the tuning for the tilt angle \(\varphi\)
was strongly related to the kernels' orientation \(\phi\),
with a circular-circular correlation coefficient \(\rho_{cc}=-0.981\)
(calculated following \citet{jammalamadaka_topics_2001}, using CircStat \cite{berens_circstat_2009}).
A scatterplot of \(\phi\) against \(\varphi\) is displayed in Fig.~\ref{fig:GaborRotVsTiltSelectivity}.
\citet{fleming_specular_2004} showed
that a set of Gabor-filters can be used to infer surface orientation in monocular images.
In images of slanted, textured surfaces,
spatial frequencies that are oriented perpendicular to the tilt angle of the surface are overrepresented.
This is due to the homographic projection on the retina,
which causes an anisotropic compression of surface textures.
The finding qualitatively extends the set of patterns that can be inferred from the LCA representation.
It adds information to the inference of surface orientation
that is different from the inference based on the local distribution of disparities.

\section{Discussion}
We add evidence to an existent body of literature,
which shows that Gabor-like, disparity tuned, phase- and position-shifted receptive fields
are a good basis for stereo algorithms (see Sec.~\ref{sec:stereoICA} and~\ref{sec:stereoBiology}).
Simple readout of disparity was possible,
due to some degree of selectivity to the stimulus
of units from the LCA representation.
These units therefore resembled Barlow's \emph{cardinal cells},
with intermediate selectivity for the stimulus.
Indeed, we did not observe a single kernel
that was not tuned for disparity or surface orientation.
At the same time, they represented a variety of other stimulus aspects,
like spatial frequency, orientation, or blob-like structures.
In combination, the kernels represented the input space well
and allowed for accurate inference.

\subsection{Dimensionality of representations}
Larger dimensionality of a representation
extended the range of disparities that could be inferred with simple readout.
We offer an intuitive explanation for this finding.
Local structure in both half-images
that originate from the same location in the world
can either be represented by one binocular kernel
with similar left and right shape;
or it can be represented by two kernels: a left and a right monocular kernel.
If binocular kernels are available,
sparsity can increase substantially,
by activation of only half the number of units
that would be needed for reconstruction with monocular kernels.
However, a representation that contains binocular kernels
requires a much larger dimensionality.

Lets assume that we want to create a new binocular representation
from \(2 \, n\) monocular kernels,
which consists of copies of the same set of \(n\) monocular half-kernels for each eye.
We generate binocular kernels from all monocular kernels,
so that the left half-kernels are shifted versions of the right half-kernels.
If we assume equally spaced horizontal and vertical shifts
in the range \(\|\Delta \bm{d}_{x,y}\|\),
the new representation has \(\sim n \, c \, \pi/4 \, \|\Delta \bm{d}_{x,y}\|^{2}\) kernels,
with factor \(c\) that determines the resolution.
Because occlusions are characterized by the lack of corresponding structure,
we would have to add the original monocular kernels to the representation,
so that the total number of kernels would be \(\sim n \, (2 \,+\, c \, \pi/4 \, \|\Delta \bm{d}_{x,y}\|^{2})\).
With either set of kernels,
the optimization can reconstruct the image equally well,
but much more sparsely with the larger set.
Obviously, the amount of information is the same in both representations,
because the information preservation constraint of the optimization is not affected.
However, the larger representation is much more redundant.
Interestingly, it is exactly this redundancy that allows for inference,
because binocular kernels that fit corresponding features are tuned for disparities.

With large dimensionality,
it was possible to infer disparity with binary classification,
even though the binarization discards information (compare \citet{bobrowski_induction_2011}).
Burge and Geisler presented an opposed approach to disparity inference,
with very low dimensionality \cite{burge_optimal_2014}.
They asked which filter shapes were optimal to infer disparity
and showed that inference was possible
with the two most informative kernels.
In their model, the activity ratio of these detectors
was the crucial parameter for inference.
Information was not distributed over many binary dimensions
but encoded in the value of a few dimensions.

In biological systems,
the value may be encoded in the firing rates of neurons.
Fine grained discriminability between neural activities,
i.e., large channel capacity,
requires high firing rates.
Indeed, there are many examples where neurons encode sensory information with high firing rates.
Examples include medial superior olivary neurons,
which lock precisely to the phase of pure tones \cite{brand_precise_2002},
and the T-units in Gymnotiforms (weakly electric fish),
which lock to the phase of electrical signals
with up to almost \num{1000}~Hz \cite{scheich_coding_1973}.
Although cortical neurons
operate at low mean firing rates of about \num{4}~Hz \cite{baddeley_responses_1997},
action potential bursts are known candidates
to encode information in firing rates.
For a current review on neural coding with bursts see \cite{zeldenrust_neural_2018}.

\subsection{The trade-off between accuracy and energy efficiency}
An alternative explanation
for the finding that cortical neurons exhibit sparse activity
is energy efficiency.
The energetic cost of a neuronal population has two major contributions:
the maintenance of neurons, which limits population size,
and neuronal activity, measured by the average rate of action potentials \cite{attwell_energy_2001,lennie_cost_2003}.
Because neuronal activity is relatively costly,
an optimization that takes energy efficiency into account
results in reduced activity \cite{levy_energy_1996}.

Indeed, the two terms of the sparse coding optimization {Eq.~\ref{eq:errorFct}}
are the preservation of information and the sparsity of the representation,
weighted against each other with the sparsity load \(\lambda\).
We have shown that the mean inference error also depends on \(\lambda\).
We therefore hypothesize that sparsity in the brain
is optimized for the trade-off
between the accuracy of upstream processing tasks and energy consumption.
This optimization could even occur dynamically and locally,
as an attention mechanism that adjusts the error subject to the current task.
Such a mechanism could interact with the error prediction we have shown,
which relies on counting the number of active coefficients.

The realization in biological substrate is plausible.
In neural notion of the LCA sparse coding,
the sparsity load corresponds to a shift in the thresholds of neurons (see Sec.~\ref{sec:LinkToBiology}).
Indeed, physiological studies show
that attentional mechanisms involve changes in the excitability of neurons.
\citet{mcadams_effects_1999} have shown
that attention modulates the response of orientation-tuned neurons in V4 multiplicatively.
Similarly, tuning maps of LCA kernels were qualitatively indifferent with respect to \(\lambda\).
Therefore, an interesting question for future research is
whether inference with variable LCA thresholds and static weights for readout is feasible.

\subsection{Model-specific issues}
Applying our processing pipeline to the naturalistic scene,
image locations with disparities
larger or smaller than the disparities included in the training set
yielded random results.
We are confident that an additional category
that includes all of the disparities beyond the included range
could successfully be added to the training set.
The category could rely on activity of coefficients of the ``Tuned Inhibitory'' type
and on lack of activity of the ``Matched Gabor'' type.
Occluded image regions are similarly characterized by the lack of corresponding image structure
and might be represented by the same kernel types.
Whether it is possible to distinguish
between a large-disparity category and an occlusion category
is an interesting question for future research.

The resolution of the disparity maps was limited in this study.
We used a stride of \num{8}\,px for the convolutional LCA sparse coding,
which was therefore also the downsampling factor for the disparity map.
With the same level of overcompleteness,
a larger stride corresponds to a larger number of kernels \cite{schultz_replicating_2014}.
We assume that a large number of kernels is mandatory
in order to represent a large number of disparities.
However, we expect that the resolution of the disparity estimates
does not depend on the stride.
Tuning maps of coefficients in a single column of the feature maps
most likely vary with respect to the position of the image structure in their receptive fields.
We may explore the limits of the spatial resolution in future research.

We have presented a naturalistic processing pipeline for disparity inference.
Our aim was not to find a method which has the lowest inference error,
but to learn more about inference based on sparse representations in general.
However, we had reasonable success of inferring disparities in a naturalistic scene.
With recent progress on neuromorphic hardware,
as well as progress on efficient implementations
of the spiking LCA algorithm \cite{zylberberg_sparse_2011,tang_sparse_2017,watkins_unsupervised_2019},
research on the hardware implementation
of our biologically inspired stereo vision processing stream
would be promising and is within reach.

\subsection{Sparse coding and supervision}
Because more overcompleteness extends the set of patterns that can be inferred,
the set of patterns
that is ecologically relevant
may predict the extend of the neuronal population in animals.
Patterns, too rare to be represented,
subject to the cost of neuronal maintenance,
should be omitted.
The likelihood that sparse coding represents patterns explicitly
may depend on the frequency of their occurrence.
The distribution might be divergent
from the relevance of the patterns an animal needs to detect.
Common patterns may be irrelevant while rare patterns,
like cues that reveal the attack of a lurking predator,
are essential for survival.
In the case of depth inference,
a uniform accuracy over the whole disparity range
might be optimal.
An advantageous learning strategy
could profit from the generality of feature extraction
based on sensory statistics
and augment learning with mild supervision
in order to gently shift the representation
towards a distribution
optimized for behavioral gain.
In multi-layered networks,
later stages might also benefit from incorporating sparsity constraints,
by aiding the clustering towards conceptional representations.
Current research supports this assumption.
\citet{kim_deep_2018} have shown that a standard autoencoder,
augmented with lateral inhibition and top-down feedback,
develops joined representations of multimodal input data.
Hale Berry Neurons were responsive for textual, as well as for visual input.
The representation was easily separable
and robust for classification tasks.

\subsection{The link between image statistics and inference}
It remains an open question
why a method
that extracts statistical properties from natural images
yields good features for inference.
The original perspective
on the independent component analysis (ICA),
a class of algorithms to which sparse coding belongs,
might point towards a possible explanation.
The reasoning behind ICA was
that data from sensor arrays
are in some cases the weighted superposition
of a number of individual, independent source signals \cite{hyvarinen_independent_2000}.
If the superposition is linear,
source signals can be reconstructed
by multiplying the vector of sensory data
with the inverse of the weight matrix.
The aim of ICA is to find this inverse matrix.

Clearly, the assumption that sensory data are the weighted sum of source signals
is not true for the formation of two-dimensional images on the retina.
Images originate from light rays scattered by objects within a physical, three-dimensional world.
The components obtained by ICA
are in fact not independent of each other \cite{bethge_factorial_2006,eichhorn_natural_2009}.
They are not the building blocks of an image
and the task of inferring depth
is not readily solved by extracting these components.
However, they seem to coincide with physical causes.
The distance of objects
manifests in the shift of corresponding image structure,
occlusions manifest in the lack of corresponding image structure,
and surface orientation manifests in anisotropically compressed texture.
Obviously, even though the feature dimensions are not the original components of the image,
they are closely linked to the geometrical layout of the scene
and therefore allow to infer properties of the external world.
They might pose the basis for a heuristic mental model of the external world,
established by the clustering of ``suspicious coincidences''
\cite{barlow_cerebral_1987,foldiak_forming_1990}.

We believe that the selectivity for patterns
that are linked to physical causes
is a general property of sparse representations of sensory data.
For example,
we have recently shown
that applying sparse coding to optic flow data
yields rather unexpected kernel shapes,
which are tuned to directions of egomotion \cite{ecke_sparse_2020}.
Screening for such selectivities
can be a starting point for identifying the cues that are at the core of inference
and it can yield predictions for properties of processing in diverse biological systems.

\section{Conclusion}
With this study,
we have extended the knowledge about similarities and differences
between representations learned with stereo sparse coding
and the visual cortex.
We have also shown
that statistical properties of the visual sensory stream
can be exploited with the sparse coding algorithm
and consecutive simple readout of depth parameters.
Disparity can be inferred reasonably well,
with very good accuracy for low disparities
but with increasing error the larger the disparity.
The range of disparities that can be inferred with good accuracy
grows with overcompleteness.
More sparsity reduces the accuracy of inference.
Since neuronal activity is directly associated with energy consumption,
attentional mechanisms could optimize the trade-off
between energy efficiency and the accuracy needed for the task an animal faces.
In addition, we have shown that accuracy of the inference
can be inferred from the number of active LCA coefficients itself.
The estimate could be used as a feedback parameter to adjust the sparsity of the optimization.

We hypothesized that sparse coding transforms the sensory stream
such that an unknown subset of patterns from the external world
can be inferred by subsequent, simple readout.
After a thorough analysis of disparity inference,
we have shown that the representation also carries information
that allows to infer surface orientation.
Selectivity for this subset of patterns
is qualitatively different from disparity tuning
because it depends on the orientation of the Gabor-like kernels shapes.
We believe that sparse coding generalizes properties from the external world
and can be used to infer a much broader range of patterns
that are cues for physical causes.

\section*{Acknowledgements}
\noindent This work was carried out at the Department of
Biology of the Eberhard-Karls-University, Tübingen, Germany.
This research did not receive any specific grant 
from funding agencies in the public, commercial, or not-for-profit sectors.

\bibliography{papers,books}

\end{document}